\let\oldvec\vec
\documentclass[runningheads]{llncs}
\pdfoutput=1

\let\vec\oldvec
\usepackage{amssymb}
\usepackage{graphicx}

\usepackage{url}
\urldef{\mailsa}\path|{alfred.hofmann, ursula.barth, ingrid.haas, frank.holzwarth,|
\urldef{\mailsb}\path|anna.kramer, leonie.kunz, christine.reiss, nicole.sator,|
\urldef{\mailsc}\path|erika.siebert-cole, peter.strasser, lncs}@springer.com|

\usepackage{rotating}
\usepackage[cmex10]{amsmath}
\usepackage{multicol}
\usepackage{graphicx}
\usepackage{thmtools}
\usepackage{graphicx}
\usepackage{amsfonts}
\usepackage{bm}
\usepackage{wrapfig}
\usepackage{subfig}
\usepackage{array}
\usepackage{makeidx}
\usepackage{moreverb}
\usepackage[lined,boxed,commentsnumbered]{algorithm2e}
\usepackage{multirow}
\usepackage{array}
\usepackage{MnSymbol}
\usepackage{bbding}
\usepackage{booktabs,tabularx}
\graphicspath{{./}}
\usepackage{epstopdf}

\DeclareMathOperator*{\argmin}{arg\,min}
\DeclareGraphicsExtensions{.jpg,.png,.tiff,.pdf}

\newcommand{\qedwhite}{\hfill \ensuremath{\Box}}
\newcommand*\samethanks[1][\value{footnote}]{\footnotemark[#1]}

\usepackage[width=122mm,left=12mm,paperwidth=146mm,height=193mm,top=12mm,paperheight=217mm]{geometry}
\begin{document}
\pagestyle{headings}
\mainmatter

\def\ECCV14SubNumber{1558}  

\title{A Graph Theoretic Approach for Object Shape Representation in Compositional Hierarchies using a Hybrid Generative-Descriptive Model}

\titlerunning{A Graph Theoretic Approach for Object Shape Representation}

\authorrunning{Umit Rusen Aktas, Mete Ozay, Ale{\v s} Leonardis and Jeremy L. Wyatt}

\author{Umit Rusen Aktas\thanks{The first and second author contributed equally.}, Mete Ozay\samethanks[1], Ale{\v s} Leonardis and Jeremy L. Wyatt
} 
\institute{School of Computer Science, The University of Birmingham, Edgbaston, Birmingham, B15 2TT, United Kingdom.
\\ Emails: \{ u.aktas, m.ozay, a.Leonardis, j.l.wyatt \} @cs.bham.ac.uk}

\maketitle

\begin{abstract}
A graph theoretic approach is proposed for object shape representation in a hierarchical compositional architecture called Compositional Hierarchy of Parts (CHOP). In the proposed approach, vocabulary learning is performed using a hybrid generative-descriptive model. First, statistical relationships between parts are learned using a \textit{Minimum Conditional Entropy Clustering} algorithm. Then, selection of \textit{descriptive} parts is defined as a frequent subgraph discovery problem, and solved using a Minimum Description Length (MDL) principle. Finally, part compositions are constructed by compressing the internal data representation with discovered substructures. Shape representation and computational complexity properties of the proposed approach and algorithms are examined using six benchmark two-dimensional shape image datasets. Experiments show that CHOP can employ part shareability and indexing mechanisms for fast inference of part compositions using learned shape vocabularies. Additionally, CHOP provides better shape retrieval performance than the state-of-the-art shape retrieval methods.

\end{abstract}


%

\section{Introduction}

\begin{figure}[!htbp]
	\centering
	\includegraphics[scale=0.195]{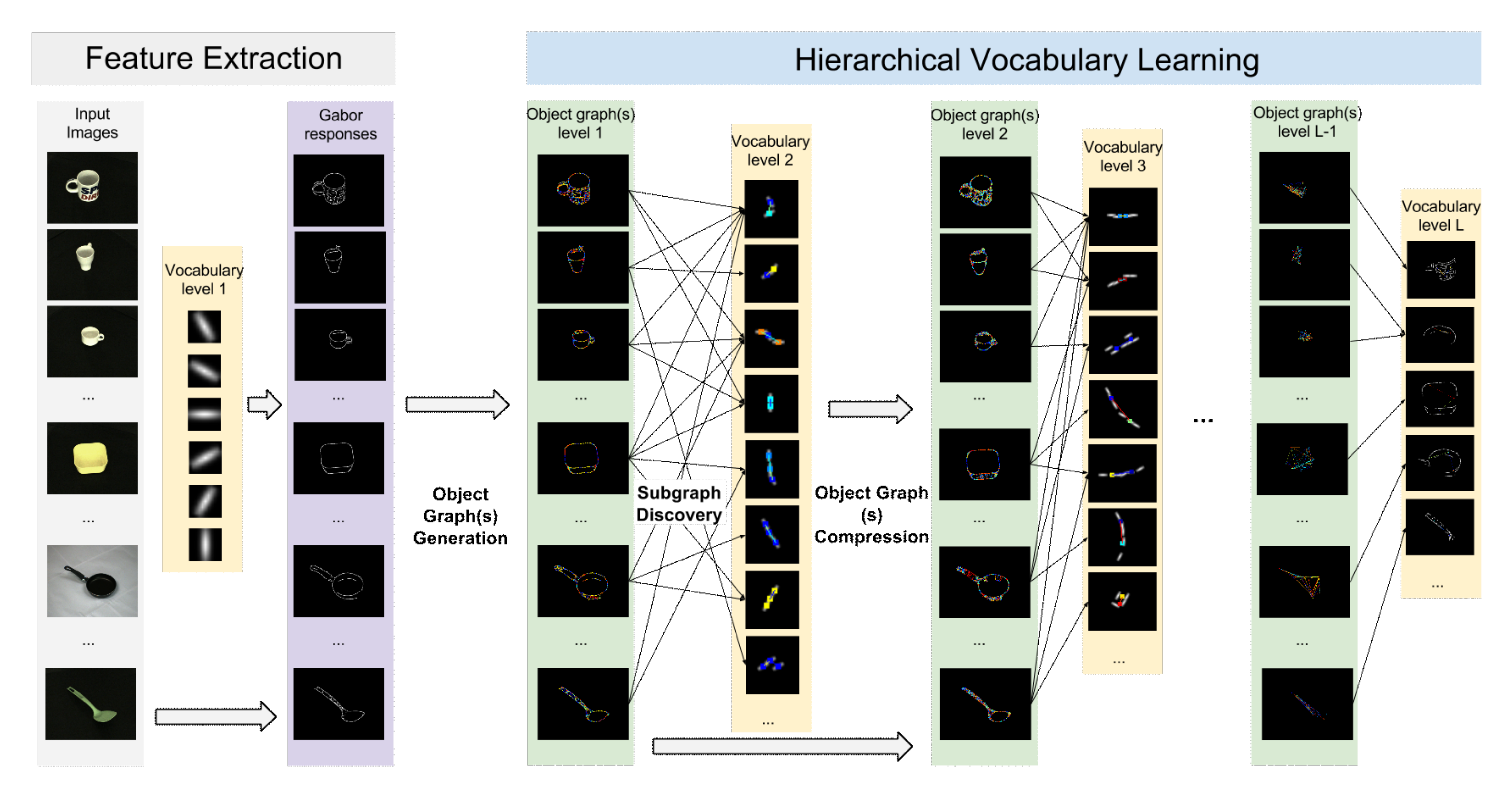}
	\caption{Overview of Compositional Hierarchy of Parts (CHOP) algorithm.}
	\label{fig:chop}
\end{figure}

Hierarchical compositional architectures have been studied in the literature as representations for object detection \cite{felz}, categorization \cite{fidler_cvpr07,comp,hdm} and parsing \cite{parsing}. A detailed review of the recent works is given in \cite{rev1}. In this paper, we propose a graph theoretic approach for object shape representation in a hierarchical compositional architecture, called Compositional Hierarchy of Parts (CHOP), using a hybrid generative-descriptive model. CHOP enables us to measure and employ generative and descriptive properties of parts for the inference of part compositions in a graph theoretic framework considering part shareability, indexing and matching mechanisms. We learn a compositional vocabulary of shape parts considering not just their statistical relationships but also their \textit{shape description} properties to generate object shapes. In addition, we take advantage of integrated models for utilization of part shareability in order to construct \textit{dense} representations of shapes in learned vocabularies for fast indexing and matching. 

A diagram demonstrating the overall structure of learning in CHOP is given in Fig. \ref{fig:chop}. At the first layer $l=1$ of CHOP, we extract Gabor features from a given set of images (Feature Extraction). We employ non-maxima suppression among Gabor feature maps in order to get local response peaks. We define parts as random graphs and represent part realizations as the instances of random graphs observed on in some dataset. At each consecutive layer, $l \geq 1$, we first learn the statistical relationships between parts using a \textit{Minimum Conditional Entropy Clustering} (MCEC) algorithm \cite{mec1} measuring conditional distributions of part realizations. For this purpose, we compute the statistical relationship between two parts $P_i$ and $P_j$ by their measuring the co-occurence statistics, for all parts represented in a learned vocabulary, and for all realizations observed on images. Using the learned statistical and spatial relationships, we encode the input data in \textit{object graphs}, where the nodes are part realizations, and edges encode discrete pairwise spatial relations (Object Graphs Generation). Next, we obtain compositions (parts) of the next layer by solving a frequent subgraph discovery problem. Each candidate subgraph (composition) is evaluated based on its ability to compress the object graphs, according to the Minimum Description Length (MDL) principle (Subgraph Discovery). Finally, part realizations for the next layer are located by compressing the object graphs using the discovered structures (Object Graphs Compression). The steps are recursively employed until no new compositions are inferred.

The paper is organised as follows. Related work and the contributions of the paper is summarized in the next section. The proposed Compositional Hierarchy of Parts (CHOP) algorithm is given in  Section \ref{sec:chop}. Experimental analyses are given in Section \ref{sec:exp}, and Section \ref{sec:conclusion} concludes the paper.

\section{Related Work and Contribution}

In \cite{deform} and \cite{l_shape}, shape models are learned using hierarchical shape matching algorithms. Kokkinos and Yuille \cite{kuki} first decompose object categories into parts and shape contours using a top-down approach. Then, they employ a Multiple Instance Learning algorithm to discriminatively learn the shape models using a bottom-up approach. However, part-shareability and indexing mechanisms \cite{lhop_book} are not employed and considered as future work in \cite{kuki}. Fidler, Boben and Leonardis \cite{lhop_book} analyzed crucial properties of hierarchical compositional approaches that should be invoked by the proposed architectures. Following their analyses, we develop an unsupervised generative-descriptive model for learning a vocabulary of parts considering part-shareability, and performing \textit{efficient} inference of object shapes on test images using an indexing and matching method. 

Fidler and Leonardis proposed a hierarchical architecture, called \textit{Learned Hierarchy of Parts} (LHOP), for compositional representation of parts \cite{fidler_cvpr07}. The main difference between LHOP and the proposed CHOP is that CHOP employs a hybrid generative-descriptive model for learning shape vocabularies using information theoretic methods in a graph theoretic framework. Specifically, CHOP first learns statistical relationships between varying number of parts, i.e. compositions of $K$-parts instead of the two-part compositions called (duplets) used in LHOP \cite{fidler_cvpr07,lhop_book}. Second, shape descriptive properties of parts are integrated with their statistical properties for inference of part compositions. In addition, the number of layers in the hierarchy are not pre-defined but determined in CHOP according to the statistical properties of the data.

MDL models have been employed for statistical shape analysis \cite{mdl1,mdl2}, specifically to achieve compactness, specificity and generalization ability properties of shape models \cite{mdl1} and segmentation algorithms \cite{boykov}. We employ MDL for the discovery of compositions of shape parts considering the statistical relationships between the parts, recursively in a hierarchical architecture. Hybrid generative-descriptive models have been used in \cite{zhu1} by employing Markov Random Fields and component analysis algorithms to construct descriptive and generative models, respectively. Although their proposed approach is hierarchical, they do not learn compositional vocabularies of  parts for shape representation.

Although our primary motivation is constructing a hierarchical compositional model for shape representation, we also examined the proposed algorithms for shape retrieval in the Experiments section. For this purpose, we compare the similarity between shapes using discriminative information about shape structures extracted from a learned vocabulary of parts and their realizations. Theoretical and experimental results of \cite{raviv,Siddiqi,Siddiqi2} on spectral properties of isomorphic graphs show that the eigenvalues of the adjacency matrices of two isomorphic graphs are ordered in an interval, and therefore provide useful information for discrimination of graphs. Assuming that shapes of the objects belonging to a category are represented (\textit{approximately}) by isomorphic graphs, we can obtain discriminative information about the shape structures by analyzing spectral properties of the part realizations detected on the shapes. \\


Our contributions in this work are threefold:
\begin{enumerate}
\item We introduce a graph theoretic approach to represent objects and parts in compositional hierarchies. Unlike other hierarchical methods \cite{felz,kuki,parsing}, CHOP learns shape vocabularies using a hybrid generative-descriptive model within a graph-based hierarchical compositional framework. The proposed approach uses graph theoretic tools to analyze, measure and employ geometric and statistical properties of parts to infer part compositions.
  
\item Two information theoretic methods are employed in the proposed CHOP algorithm to learn the statistical properties of parts, and construct compositions of parts. First we learn the relationship between parts using MCEC \cite{mec1}. Then, we select and infer compositions of parts according to their shape description properties defined by an MDL model.

\item CHOP employs a hybrid generative-descriptive model for hierarchical compositional representation of shapes. The proposed model differs from frequency-based approaches in that the part selection process is driven by the MDL principle, which effectively selects parts that are both frequently observed and provide \textit{descriptive} information for the representation of shapes. 
\end{enumerate}

\section{Compositional Hierarchy of Parts}
\label{sec:chop}

In this section, we give the descriptions of the algorithms employed in CHOP in its training and testing phases. In the next section, we first describe the preprocessing algorithms that are used in both training and testing. Next, we introduce the vocabulary learning algorithms in Section \ref{sec:lvp}. Then, we describe the inference algorithms performed on the test images in Section \ref{sec:test}.

\subsection{Preprocessing}
\label{sec:pp}

Given a set of images $S= \{ s_n, y_n \}_{n=1}^N$, where $y_n \in \mathbb{Z}^{+}$ is the category label of an image $s_n$, we first extract a set of Gabor features $F_n=\{ f_{nm} (\mathbf{x}_{nm}) \in \mathbb{R} \}_{m=1}^M$ from each image $s_n$ using Gabor filters employed at location $\mathbf{x}_{nm}$ in $s_n$ at $\Theta$ orientations \cite{fidler_cvpr07}. Then, we construct a set of Gabor features $F=\bigcup \limits _{n=1} ^N F_n$. In this work, we compute the Gabor features at $\Theta=6$ different orientations. In order to remove the redundancy of Gabor features, we perform non-maxima suppression. In this step, a Gabor feature with the Gabor response value $f_{nm} (\mathbf{x}_{nm})$ is removed from $F_n$ if $f_{nm} (\mathbf{x}_{nm}) < f_{na} (\mathbf{x}_{na})$, for all Gabor features extracted at $\mathbf{x}_{na} \in \aleph (\mathbf{x}_{nm})$, where $\aleph (\mathbf{x}_{nm})$ is a set of image positions of the Gabor features that reside in the neighborhood of $\mathbf{x}_{nm}$ defined by Euclidean distance in $\mathbb{R}^2$. Finally, we obtain a set of suppressed Gabor features $\hat{F}_n \subset F_n$ and  $\hat{F}=\bigcup \limits _{n=1} ^N \hat{F}_n$.


\subsection{Learning a Vocabulary of Parts}
\label{sec:lvp}

Given a set of training images $S^{tr}$, we first learn the statistical properties of parts using their realizations on images at a layer $l$. Then, we infer the compositions of parts at layer $l+1$ by minimizing the description length of the object descriptions defined as \textit{Object Graphs}. In order to remove the redundancy of the compositions, we employ a \textit{local inhibition} process that was suggested in \cite{fidler_cvpr07}. Statistical learning of part structures, inference of compositions and local inhibition processes are performed by constructing compositions of parts at each layer, recursively, and the details are given in the following subsections. 
 
\begin{definition}[Parts and Part Realizations]
\label{def:part}

The $i^{th}$ part constructed at the $l^{th}$ layer $\mathcal{P}_i ^l=(\mathcal{G}_i ^l, \mathcal{Y}_i^l)$ is a tuple consisting of a directed random graph $\mathcal{G}_i ^l= ( \mathcal{V}_i^l, \mathcal{E}_i^l )$, where $\mathcal{V}_i^l$ is a set of nodes and $\mathcal{E}_i^l$ is a set of edges, and $\mathcal{Y}_i^l \in \mathbb{Z}^+$ is a random variable which represents the identity number or label of the part. The realization $R_i^l(s_n)=(G_i ^l(s_n), Y_i^l(s_n))$  of $\mathcal{P}_i ^l$ is defined by $1)$ $Y_i^l(s_n)$ which is the realization of $\mathcal{Y}_i^l $ representing the label of the part realization on an image $(s_n)$, and $2)$ the directed graph $G_i ^l(s_n)= \{ V_i^l(s_n) , E_i^l(s_n) \}$ which is an instance of the random graph $\mathcal{G}_i ^l$ computed on a training image $(s_n) \in  S^{tr}$, where $V_i^l(s_n)$ is a set of nodes and $E_i^l(s_n)$ is a set of edges of $G_i ^l(s_n)$, $\forall n=1,2, \ldots, N_{tr}$. 

At the first layer $l=1$, each node of $\mathcal{V}_i ^1$ is a part label $\mathcal{Y}_i^1 \in  \mathcal{V}_i ^1$ taking values from the set $\{ 1, 2, \ldots, \Theta \}$, and $\mathcal{E}_i ^1 = \emptyset$. Similarly, $E_i ^1(s_n) = \emptyset$, and each node of $V_i^1(s_n)$ is defined as a Gabor feature $f^i_{na} (\mathbf{x}_{na}) \in \hat{F}_n ^{tr}$ observed in the image $s_n \in S^{tr}$ at the image location $\mathbf{x}_{na}$, i.e. the $a^{th}$ realization of $\mathcal{P}_i ^l$ observed in $s_n \in S^{tr}$ at $\mathbf{x}_{na}$, $\forall n=1,2,\ldots,N_{tr}$. In the consecutive layers, the parts and part realizations are defined recursively by employing layer-wise mappings $\Psi_{l,l+1}$ defined as
\begin{equation}
\Psi_{l,l+1}: (\mathcal{P} ^l, R ^l,\mathbb{G}_l) \to (\mathcal{P}^{l+1}, R^{l+1}), \forall l=1,2, \ldots, L,
\label{eq:map}
 \end{equation} 
where $\mathcal{P} ^l= \{ \mathcal{P}_i ^l \}_{i=1} ^{A_l}$, ${R} ^l= \{ {R}_i ^l (s_n): s_n \in S^{tr} \}_{i=1} ^{B_l}$, $\mathcal{P} ^{l+1}= \{ \mathcal{P}_j ^{l+1} \}_{j=1} ^{A_{l+1}}$, ${R} ^{l+1}= \{ {R}_j ^{l+1} (s_n): s_n \in S^{tr} \}_{j=1} ^{B_{l+1}}$ and $\mathbb{G}_l$ is an object graph which is defined next.
\qedwhite
\end{definition}

In the rest of this section, we will use ${R}_j ^l (s_n) \triangleq {R}_j ^l$, $\forall j=1,2,\ldots,B_l$, $\forall l=1,2, \ldots, L$, $\forall s_n \in S^{tr}$, for the sake of simplicity in the notation.

\begin{definition}[Receptive and Object Graph]
\label{def:receptive_field}

A receptive graph of a part realization $\mathcal{R}_i ^{l}$ is a star-shaped graph ${RG_i ^{l} = (V_i^{l}, E_i^{l})}$, which is induced from a receptive field centered at the root node $\mathcal{R}_i ^{l}$. A directed edge $e_{ab} \in E_i^{l}$ is defined as
\begin{equation}
e_{ab}= 
\begin{cases}
(a^l, b^l, \phi_{ab}^l), & \quad {\rm if} \quad \mathbf{x}_{nb} \in \aleph (\mathbf{x}_{na}), a=i \ \\
\emptyset, & \quad otherwise
\end{cases} ,
\label{eq:mdl_edges}
\end{equation} 
where $\aleph (\mathbf{x}_{na})$ is the set of part realizations that reside in a neighborhood of a part realization $R_a^l$ in an image $s_n$, $\forall{R_a^l,R_b^l} \in V^l_i, b \neq i$ and $\forall s_n \in S^{tr}$. $\phi_{ab}^l$ defines the statistical relationship between $R_a^l$ and $R_b^l$, as explained in the next subsection.

The structure of part realizations observed at the $l^{th}$ layer on the training set $S^{tr}$ is \textit{described} using a directed graph $\mathbb{G}_l=(\mathbb{V}_l, \mathbb{E}_l)$, called an object graph, where $\mathbb{V}_l = \bigcup \limits _i V_i^{l}$ is a set of nodes, and $\mathbb{E}_l = \bigcup \limits _i E_i^{l}$ is a set of edges, where $V_i$ and $E_i$ is the set of nodes and edges of a receptive graph $RG_i$, $\forall i$, respectively.
\qedwhite
\end{definition}

\textbf{Learning of Statistical Relationships between Parts and Part Realizations} We compute the \textit{conditional} distributions $P _{\mathcal{P}_i^l} (R_a^l| \mathcal{P}_j^l=R_b^l)$ for each $i=Y_a^l$ and $j=Y_b^l$ between all possible pairs of parts $(\mathcal{P}_i^l,\mathcal{P}_j^l)$ using $S^{tr}$ at the $l^{th}$ layer. However, we select a set of modes $\mathcal{M}^l= \{ M_{ij}: i=1,2, \ldots, B_l, j=1,2, \ldots, B_l \}$, where $M_{ij}= \{ M_{ijk}  \} _{k=1} ^K$ of these distributions instead of detecting a single mode. For this purpose, we define the mode computation problem as a \textit{Minimum Conditional Entropy Clustering} problem \cite{mec1} as
\begin{equation}
Z_{ijk}:=\argmin _{\pi_k \in C} H(\pi_k, R_a^l| R_b^l), 
\label{eq:cond_entropy_min}
\end{equation} 
\begin{equation}
H(\pi_k, R_a^l| R_b^l)= - \sum _{ \forall \mathbf{x}_{na} ^l \in \aleph ( \mathbf{x}_{nb}^l ) } \sum \limits _{k=1} ^{K} P (\pi_k, R_a^l| R_b^l) \log P (\pi_k, R_a^l| R_b^l). 
\label{eq:cond_entropy}
\end{equation} 
The first summation is over all part realizations $R_a^l$ that reside in a neighborhood of all $R_b^l$ such that $ \mathbf{x}_{na} ^l \in \aleph ( \mathbf{x}_{nb}^l )$, for all $i=Y_a^l$ and $j=Y_b^l$, $C$ is a set of cluster ids, $K=|C|$ is the number of clusters, $\pi_k \in C$ is a cluster label, and $P (\pi_k, R_a^l| R_b^l) \triangleq P _{\mathcal{P}_i^l} (\pi_k,R_a^l| \mathcal{P}_j^l=R_b^l) $. 

The pairwise statistical relationship between two part realizations $R_a^l$ and $R_b^l$ is represented as $M_{ijk}= ( i, j, \mathbf{c}_{ijk},Z_{ijk} )$, where $\mathbf{c}_{ijk}$ is the center position of the $k^{th}$ cluster. In the construction of an object graph $\mathbb{G}_l$ at the $l^{th}$ layer, we compute $\phi_{ab}^l= (\mathbf{c}_{ijk},\hat{k})$, $\forall a,b$ as $ \hat{k}=\argmin _{k \in C} \|  \mathbf{d}_{ab} - \mathbf{c}_{ijk} \|_2$, where $\| \cdot \|_2$ is the Euclidean distance, $i=Y_a^l$ and $j=Y_b^l$, $\mathbf{d}_{ab} = \mathbf{x}_{na} - \mathbf{x}_{nb}$, $\mathbf{x}_{na}$ and $\mathbf{x}_{nb}$  are the positions of $R_a^l$ and $R_j^l$ in an image $s_n$, respectively.

\textbf{Inference of Compositions of Parts using MDL}  Given a set of parts $\mathcal{P} ^l$, a set of part realizations $\mathcal{R} ^l$, and an object graph $\mathbb{G}_l$ at the $l^{th}$ layer, we infer compositions of parts at the $(l+1)^{st}$ layer by computing a mapping $\Psi_{l,l+1}$ in \eqref{eq:map}. In this mapping, we search for a structure which \textit{best describes} the structure of parts $\mathcal{P} ^l$ as the compositions constructed at the $(l+1)^{st}$ layer by minimizing the length of description of $\mathcal{P} ^l$. In the inference process, we search a set of graphs $\mathcal{G} ^{l+1} = \{ \mathcal{G}_j ^{l+1} \}_{j=1} ^{A_{l+1}}$ which minimizes the description length of $\mathbb{G}_l$ as

\begin{equation}
{\mathcal{G}} ^{l+1}= \argmin _{ \mathcal{G}_j ^{l+1} } value(\mathcal{G}_j ^{l+1}, \mathbb{G}_l),
\label{eq:dl}
\end{equation}
where
\begin{equation}
value(\mathcal{G}_j ^{l+1}, \mathbb{G}_l)= \frac{DL(\mathcal{G}_j ^{l+1})+DL(\mathbb{G}_l|\mathcal{G}_j ^{l+1})}{DL(\mathbb{G}_l)}.
\label{eq:val}
\end{equation}
is the compression value of an object graph $\mathbb{G}_l$ given a subgraph $\mathcal{G}_j ^{l+1}$ of a receptive graph ${RG}_j^{l}$, $\forall j=1,2,\ldots,{B_{l}}$. Description length $DL$ of a graph $G$ is calculated using the number of bits to represent node labels, edge labels and adjacency matrix, as explained in \cite{ref:Subdue}. The inference process consists of two steps:

\begin{figure}[t]
	\centering
	\subfloat[$\mathcal{G}_1^{l+1}$ (Valid)]{\includegraphics[scale=0.2]{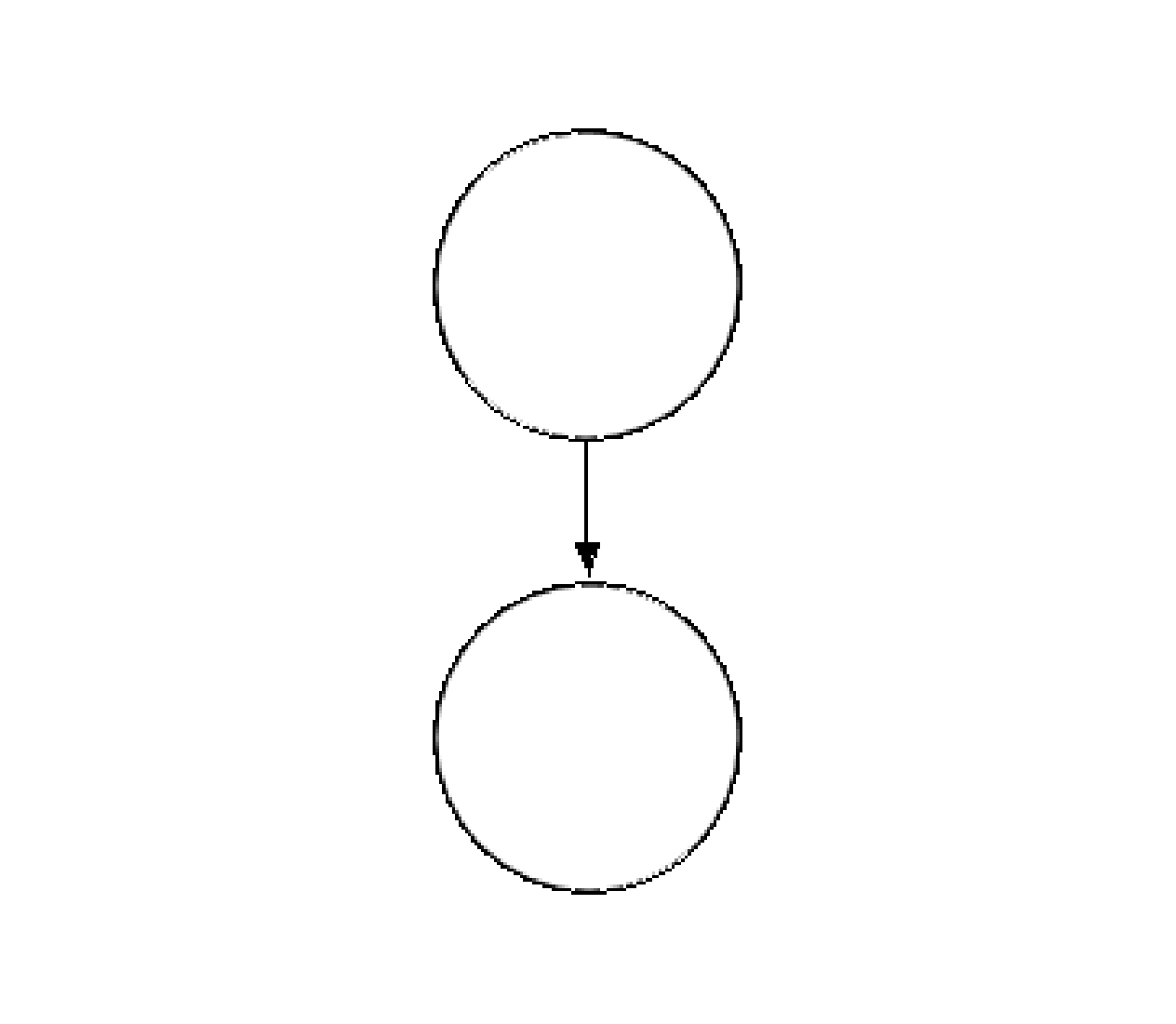}} \hspace{0.1cm}
	\subfloat[$\mathcal{G}_2^{l+1}$ (Valid)]{\includegraphics[scale=0.2]{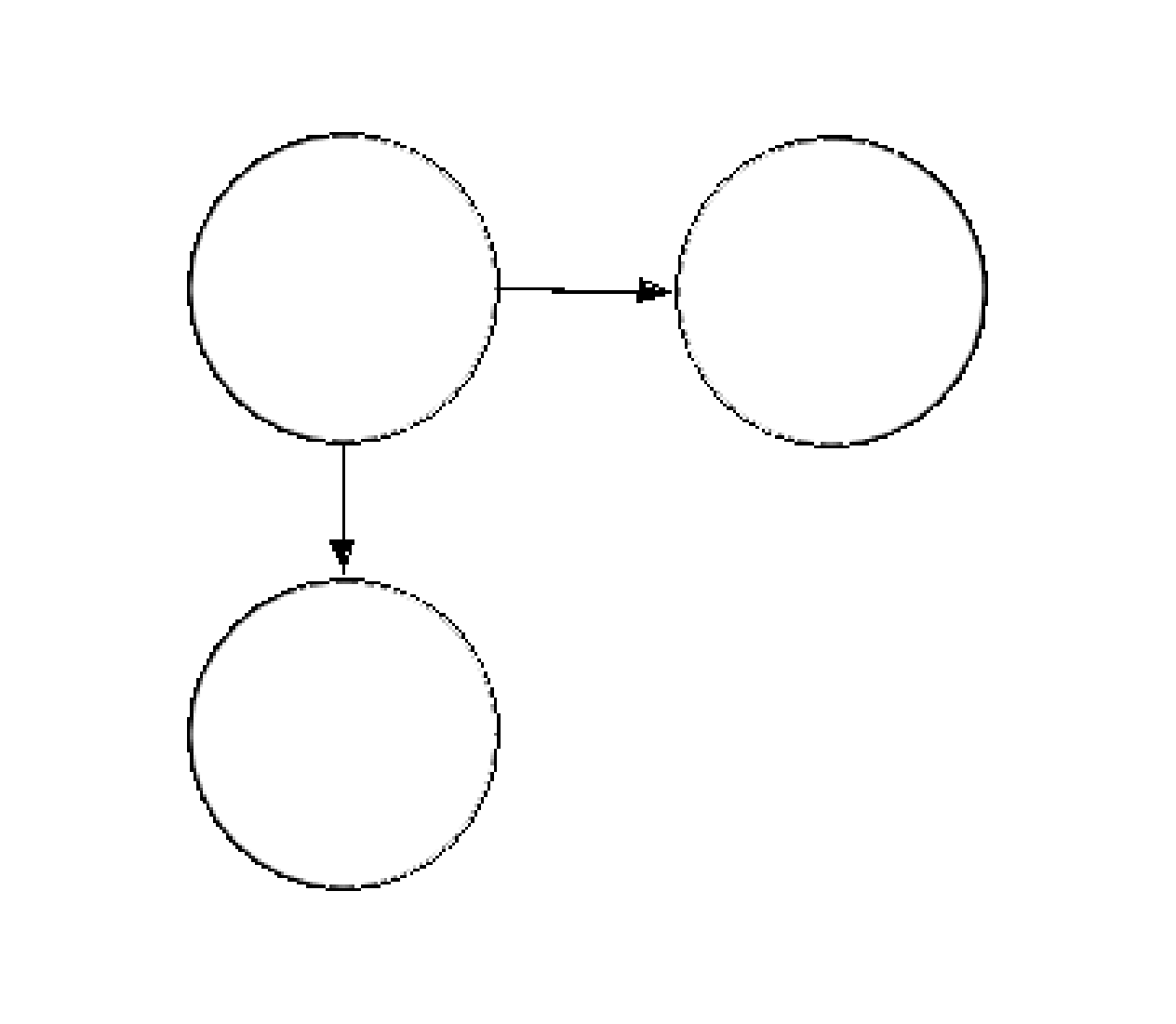}} \hspace{0.1cm}
	\subfloat[$\mathcal{G}_3^{l+1}$ (Invalid)]{\includegraphics[scale=0.2]{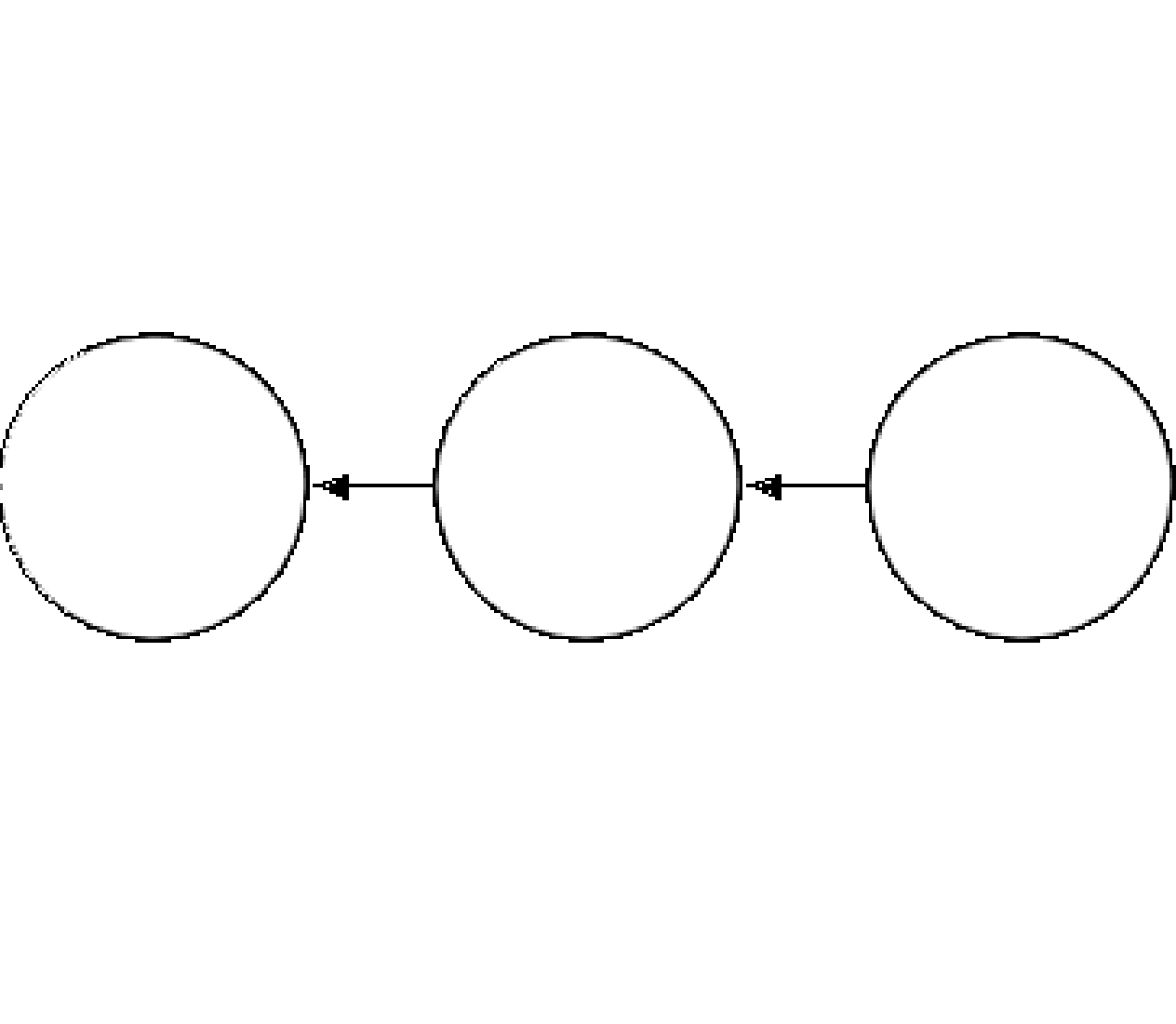}} \hspace{0.1cm}
	\subfloat[$\mathcal{G}_4^{l+1}$ (Invalid)]{\includegraphics[scale=0.2]{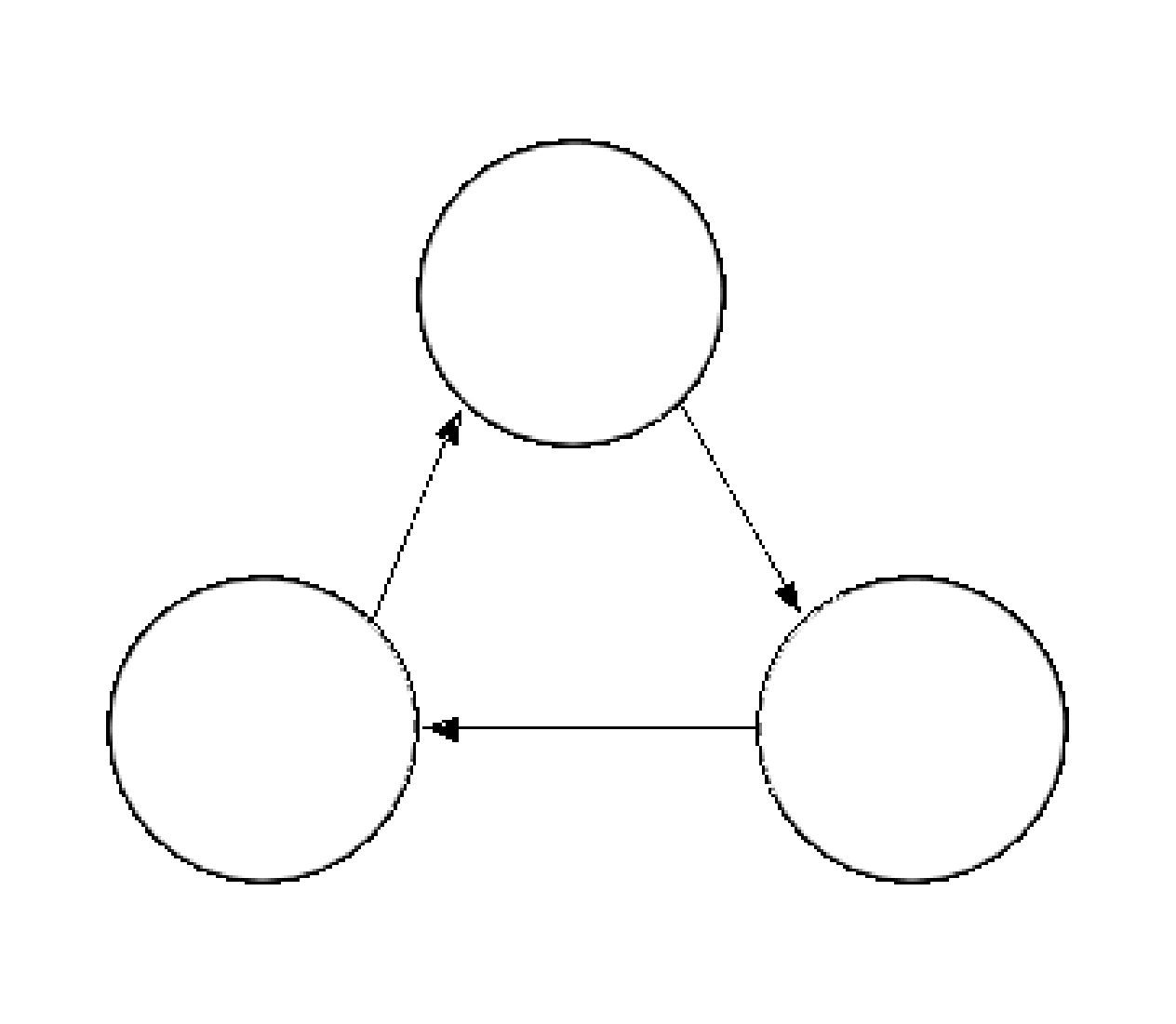}} \hspace{0.1cm}
	\caption{Valid and invalid candidates.}
	\label{fig:candidates}
\end{figure}
\newpage
\begin{enumerate}
\item \textbf{Enumeration:} In the graph enumeration step, candidate graphs $\mathcal{G}^{l+1}$ are generated from $\mathbb{G}_l$. However, each $\mathcal{G}_j^{l+1} \in \mathbb{G}_l$ is required to include nodes $\mathcal{V}_j^{l+1}$ and edges $\mathcal{E}_j^{l+1}$ from only one receptive graph $RG_i ^{l}$, $\forall i$. This selective candidate generation procedure enforces $\mathcal{G}_j^{l+1}$ to represent an area around its centre node. Examples of valid and invalid candidates are illustrated in Fig. \ref{fig:candidates}. $\mathcal{G}_1^{l+1}$ and $\mathcal{G}_2^{l+1}$ are valid structures since each graph is inferred from a single receptive graph, e.g. $RG_1 ^{l}$  and $RG_2 ^{l}$, respectively. Invalid graphs $\mathcal{G}_3^{l+1}$ and $\mathcal{G}_4^{l+1}$ are not enumerated since their nodes/edges are inferred from multiple receptive graphs.

\item \textbf{Evaluation:} Once we obtain $ \mathcal{G} ^{l+1}$ by solving \eqref{eq:dl} with $ \mathcal{G} ^{l+1}$ subject to constraints provided in the previous step, we compute a set of graph instances of part realizations $G ^{l+1}= \{ G_i ^{l+1} \} _{i=1} ^{ B_{l+1} }$ such that $G_i ^{l+1} \in iso(\mathcal{G}_j ^{l+1})$ and $G_i ^{l+1} \subseteq \mathbb{G}_l$, where $iso( \mathcal{G}_j ^{l+1})$ is a set of all subgraphs that are isomorphic to $\mathcal{G}_j ^{l+1}$. This problem is defined as a subgraph isomorphism problem \cite{subdue}, which is NP-complete. In this work, the proposed graph structures are acyclic and star-shaped, enabling us to solve \eqref{eq:dl} in P-time. In order to obtain two sets of subgraphs $\mathcal{G} ^{l+1}$ and $G ^{l+1}$ by solving \eqref{eq:dl}, we have implemented a simplified version of the substructure discovery system, SUBDUE \cite{subdue} which is employed in a restricted search space. The discovery algorithm is explained in Algorithm \ref{alg:subdue}. The key difference between the original SUBDUE and our implementation is that in Step $\mathbf{4}$, $childList$ contains only star-shaped graphs, which are extended from $parentList$ by single nodes. The parameters $beam$, $numBest$, $bestPartSize$ are used to prune the search space. 
\end{enumerate}

The label of a part $\mathcal{P}_j ^{l+1}$ is defined according to its compression value $ \mu_j ^{l+1} \triangleq value(\mathcal{G}_j ^{l+1}, \mathbb{G}_l)$ computed in \eqref{eq:val}. We sort compression values in ascending order, and assign the part label to the index of the compression value of the part.

After sets of graphs and part labels are obtained at the $(l+1)^{st}$ layer, we construct a set of parts $\mathcal{P} ^{l+1}= \{ \mathcal{P}_i ^{l+1} \} _{i=1} ^{A_{l+1}}$, where $ \mathcal{P}_i ^{l+1}=(\mathcal{G}_i ^{l+1},\mathcal{Y}_i^{l+1})$. We call $\mathcal{P} ^{l+1}$ a set of \textit{compositions} of the parts from $\mathcal{P} ^{l}$, constructed at the $(l+1)^{st}$ layer. Similarly, we extract a set of part realizations $\hat{R} ^{l+1}= \{ {R}_j ^{l+1} \} _{j=1} ^{B_{l+1}}$, where ${R}_j ^{l+1}= ( G_j ^{l+1},Y_j^{l+1} )$. 
In order to remove the redundancy in $\hat{R} ^{l+1}$, we perform local inhibition as in \cite{fidler_cvpr07} and obtain a new set of part realizations ${R} ^{l+1} \subseteq \hat{R} ^{l+1}$.\\
\vspace{0.5cm}

\begin{algorithm}[t!]
\SetKwFunction{Union}{Union}\SetKwFunction{FindCompress}{FindCompress}
\SetKwInOut{Input}{Input}\SetKwInOut{Output}{Output}
\Input { $\mathbb{G}_l=(\mathbb{V}_l, \mathbb{E}_l)$: Object graph, $beam$, $numBest$, $bestPartSize$.
}
\Output{ Parts $\mathcal{P}^{l+1}$, realizations $\mathcal{R}^{l+1}$.
}	
	\nl $parentList:=null$; $childList:=null$; $bestPartList:=null$;\\
	where $childList$,$bestPartList$ are priority queues ordered by MDL scores.\\ 
	\nl Initialize $parentList$ with frequent single node parts;\\
	\nl\While{$parentList$ is not empty}{
		\nl Extend parts in $parentList$ in all possible ways into $childList$;\\
		\nl Evaluate parts in $childList$ using \eqref{eq:val};\\
		\nl Trim $childList$ to $beam$ top parts;\\
		\nl Merge elements of $childList$ and $bestPartList$ into $bestPartList$;\\
		\nl $parentList:=null$;\\
		\nl Swap $parentList$ and $childList$;\\
	}
	\nl Trim $bestPartList$ to $maxBest$ top parts;\\
	\nl $\mathcal{P}^{l+1}:=bestPartList$;\\
	\nl $\mathcal{R}^{l+1}:=bestPartList.getInstances()$;
\caption{Inference of new compositions.}
\label{alg:subdue}
\end{algorithm}

\textbf{Incremental Construction of the Vocabulary}
\begin{definition}[Vocabulary]
\label{def:voc}
A tuple $\Omega_l=( \mathcal{P} ^l, \mathcal{M} ^l )$ is the vocabulary constructed at the $l^{th}$ layer using the training set $S^{tr}$. The vocabulary of a CHOP with $L$ layers is defined as the set $ \Omega= \{ \Omega_l : l=1,2, \ldots, L \}$.
\qedwhite
\end{definition}

\begin{algorithm}[t!]
%
\SetKwFunction{Union}{Union}\SetKwFunction{FindCompress}{FindCompress}
\SetKwInOut{Input}{Input}\SetKwInOut{Output}{Output}
\Input {
\begin{itemize}
\item $S^{tr}= \{ s_n \} _{n=1} ^{N}$: Training dataset, 
\item $\Theta$: The number of different orientations of Gabor features,
\item $\sigma$: Subsampling ratio.
\end{itemize}
}
\Output{Vocabulary $\Omega$.}

\nl Extract a set of Gabor features $F^{tr}=\bigcup \limits _{n=1} ^N F^{tr}_n$, where $F^{tr}_n=\{ f_{nm} (\mathbf{x}_{nm}) \}_{m=1}^M$ from each image $s_n \in S^{tr}$; \\
\nl Construct a set of suppressed Gabor features $\hat{F}^{tr} \subset F^{tr}$ (see Section \ref{sec:pp}); \\
\nl $l:= 1$; \\
\nl Construct $\mathcal{P} ^1$ and ${R} ^1$ (see Definition \ref{def:part}); \\ 
\While{$\mathcal{G}^{l} \neq \emptyset$}{
\nl Compute the sets of modes $\mathcal{M}^l$ (see Section \ref{sec:lvp}); \\
\nl Construct $\mathbb{G}_l$ using $\mathcal{M}^l$ (see Definition \ref{def:receptive_field}); \\ 
\nl Construct $\Omega_l=( \mathcal{P} ^l, \mathcal{M}^l )$; \\ 
\nl Infer part graphs $\mathcal{G}^{l+1}$ by solving \eqref{eq:dl} (see Section \ref{sec:lvp});\\
\nl Construct $\mathcal{P} ^{l+1}$ and ${R} ^{l+1}$ (see Section \ref{sec:lvp});\\
\nl $l:=l+1$;\\
\nl Subsample the positions of part realizations $R^{l}_{i}$ by a factor of $\sigma$, $\forall n, R^{l}_{i}$;
}
\nl $ \Omega= \{ \Omega_t : t=1,2, \ldots, l-1 \}$;
\caption{ The vocabulary learning algorithm of Compositional Hierarchy of Parts.}
\label{alg:voc_learn}
\end{algorithm}

We construct $\Omega$ of CHOP incrementally as described in the pseudo-code of the vocabulary learning algorithm given in Algorithm \ref{alg:voc_learn}. In the first step of the algorithm, we extract a set of Gabor features $F_n=\{ f_{nm} (\mathbf{x}_{nm}) \}_{m=1}^M$ from each image $s_n \in S^{tr}$ using Gabor filters employed at location $\mathbf{x}_{nm}$ in $s_n$ at $\Theta$ orientations. Then, we perform local inhibition of Gabor features using non-maxima suppression to construct a set of suppressed Gabor features $\hat{F}_n \subset F_n$ as described in Section \ref{sec:pp} in the second step. Next, we initialize the variable $l$ which defines the layer index, and we construct parts $\mathcal{P} ^1$ and part realizations ${R} ^1$ at the first layer as described in Definition \ref{def:part}.

In steps $\mathbf{5}-\mathbf{11}$, we incrementally construct the vocabulary of the CHOP. In step $\mathbf{5}$, we compute the sets of modes $\mathcal{M}^l$ by learning statistical relationships between part realizations as described in Section \ref{sec:lvp}. In the sixth step, we construct an object graph $\mathbb{G}_l$ using $\mathcal{M}^l$ as explained in Definition \ref{def:receptive_field}, and we construct the vocabulary $\Omega_l=( \mathcal{P} ^l, \mathcal{M}^l )$ at the $l^{th}$ layer in step $\mathbf{7}$. Next, we infer part graphs that will be constructed at the next layer $\mathcal{G}^{l+1}$ by computing the mapping $\Psi_{l,l+1}$. For this purpose, we solve \eqref{eq:dl} using our graph mining implementation to obtain a set of parts $\mathcal{P} ^{l+1}$ and a set of part realizations ${R} ^{l+1}$ as explained in Section \ref{sec:lvp}. We increment $l$ in step $\mathbf{10}$, and subsample the positions of part realizations $R^{l}_{i}$ by a factor of $\sigma$, $\forall n, R^{l}_{i}$ in step $\mathbf{11}$, which effectively increases the area of the receptive fields through upper layers. We iterate the steps $\mathbf{5}-\mathbf{11}$ while a non-empty part graph $\mathcal{G}_i^{l}$ is either obtained from the training images at the first layer, or inferred from $\Omega_{l-1}$, $R^{l-1}$ and $\mathbb{G}_{l-1}$ at $l > 1$, i.e. $\mathcal{G}^{l} \neq \emptyset$, $\forall l \geq 1$. As the output of the algorithm, we obtain the vocabulary of CHOP, $ \Omega= \{ \Omega_l : l=1,2, \ldots, L \}$.

\subsection{Inference of Object Shapes on Test Images}
\label{sec:test}

\begin{figure}[!t]
	\centering
	\includegraphics[scale=0.2]{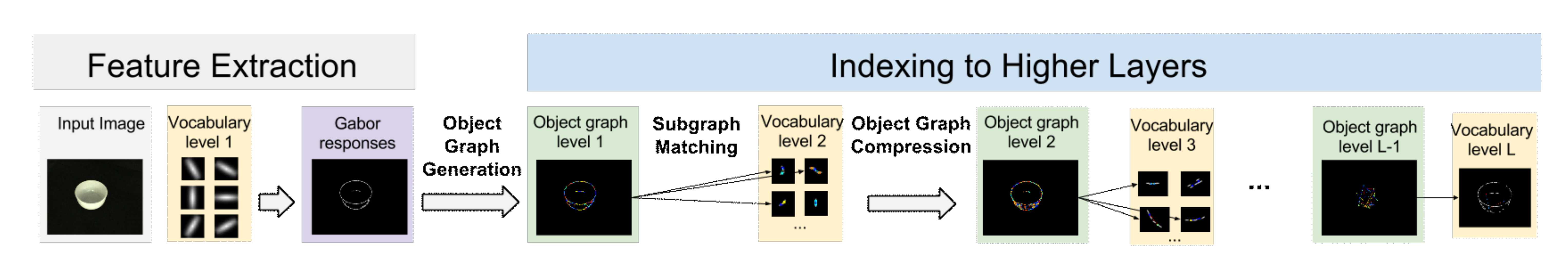}
	\caption{Inference in Compositional Hierarchy of Parts (CHOP) framework.}
	\label{fig:chopinference}
\end{figure}

In the testing phase, we infer shapes of objects on test images $s_n \in S^{te}$ using the learned vocabulary of parts $\Omega$. The algorithm flow of our inference algorithm resembles that of learning, as shown in Fig. \ref{fig:chopinference}. The only difference between the learning and inference processes is that no new subgraphs are \textit{discovered} from the input image in inference, but learned compositions are matched to their instances (Subgraph Matching). Algorithm \ref{alg:voc_test} explains the inference algorithm for test images. We incrementally construct a set of inference graphs $\mathcal{T}(s_n)= \{ \mathcal{T}_{l}(s_n) \}_{l=1} ^L$ of a given test image $s_n \in S^{te}$ using the learned vocabulary $\Omega= \{ \Omega_l \}_{l=1} ^L$. At each $l^{th}$ layer, we construct a set of part realizations $R^l(s_n)= \Big \{ R_i^l(s_n)= \Big( G_i ^l(s_n), Y_i^l(s_n) \Big) \Big \}  _{i=1} ^{{B'}_l}$ and an object graph $\mathbb{G}_l=(\mathbb{V}_l, \mathbb{E}_l)$ of $s_n$, $\forall l=1,2,\ldots,L$. The test image is processed in the same manner as in vocabulary learning (steps $\mathbf{1}-\mathbf{5}$). In step $\mathbf{6}$, isomorphisms of part graph descriptions $\mathcal{G}^{l+1}$ obtained from $\Omega_{l+1}$ are searched in $\mathbb{G}_{l}$ in P-time (see Section \ref{sec:lvp}). Part realizations ${R} ^{l+1}$ of the new object graph $\mathbb{G}_{l+1}$ are extracted from $G ^{l+1}$ in step $\mathbf{7}$. The discovery process continues until no new realizations are found.

\newpage
At the first layer $l=1$, the nodes of the instance graph $G_i ^1(s_n)$ of a part realization $R_i^1(s_n)$ represent the Gabor features $f^i_{na} (\mathbf{x}_{na}) \in \hat{F}_n ^{te}$ observed in the image $s_n \in S^{te}$ at an image location $\mathbf{x}_{na}$ as described in Section \ref{sec:lvp}. In order to infer the graph instances and compositions of part realizations in the following layers $1<l \leq L$, we employ a graph matching algorithm that constructs $G_i ^{l+1}(s_n)= \{ H(\mathcal{P} ^{l+1}): H(\mathcal{P} ^{l+1}) \subseteq \mathbb{G}_l \}$ which is a set of subgraph isomorphisms $H(\mathcal{P} ^{l+1})$ of part graphs $\mathcal{G} ^{l+1}$ in $\mathcal{P} ^{l+1}$, computed in $\mathbb{G}_l$.

\begin{algorithm}[h!]
%
\SetKwFunction{Union}{Union}\SetKwFunction{FindCompress}{FindCompress}
\SetKwInOut{Input}{Input}\SetKwInOut{Output}{Output}
\Input {
\begin{itemize}
\item $s$: Test image, 
\item $\Omega$: Vocabulary,
\item $\Theta$: The number of different orientations of Gabor features,
\item $\sigma$: Subsampling ratio.
\end{itemize}
}
\Output{Inference graph $\mathcal{T}(s)$.}

\nl Extract a set of Gabor features $F= \{ f_{m} (\mathbf{x}_{m}) \} _{m=1}^M$ from image $s$; \\
\nl Construct a set of suppressed Gabor features $\hat{F} \subset F$ (see Section \ref{sec:pp}); \\
\nl $l:= 1$; \\
\nl Construct ${R}^1$ from $\hat{F}$ (see Definition \ref{def:part}); \\ 
\While{$\Omega_{l+1} \neq \emptyset \wedge {R}^{l} \neq \emptyset$ }{
\nl Construct $\mathbb{G}_l$ using $\mathcal{M}^l$ in $\Omega_{l}$; \\
\nl Find graph instances of part realizations $G ^{l+1}= \{ G_j ^{l+1} \} _{j=1} ^{{B'}_{l+1}}$ such that $G_j ^{l+1} \in iso(\mathcal{G}^{l+1})$ and $G_j ^{l+1} \subseteq \mathbb{G}_l$ (see Section \ref{sec:lvp}, \textbf{Evaluation});\\
\nl Construct ${R} ^{l+1}$ from $G ^{l+1}$ (see Section \ref{sec:lvp});\\
\nl $l:=l+1$;\\
\nl Subsample the positions of part realizations $R^{l}_{i}$ by a factor of $\sigma$, $\forall R^{l}_{i}$;
}
\nl $ \mathcal{T}(s) = \{ \mathbb{G}_t : t=1,2, \ldots, l-1 \}$;
\caption{Object shape inference algorithm for test images.}
\label{alg:voc_test}
\end{algorithm}
\newpage
\section{Experiments}
\label{sec:exp}

We examine our proposed approach and algorithms on six benchmark object shape datasets, which are namely the Washington image dataset (Washington) \cite{aloi}, the MPEG-7 Core Experiment CE-Shape 1 dataset \cite{mpeg7}, the ETHZ Shape Classes dataset \cite{ethshape}, 40 sample articulated Tools dataset (Tools-40) \cite{tools-40}, 35 sample multi-class Tools dataset (Tools-35) \cite{tools} and the Myth dataset \cite{tools}. In the experiments, we used $\Theta=6$ different orientations of Gabor features with the same Gabor kernel parameters implemented in \cite{fidler_cvpr07}. We used a subsampling ratio of $\sigma=0.5$. A Matlab implementation of CHOP is available \href{https://github.com/rusen/CHOP.git}{here}\footnote{\url{https://github.com/rusen/CHOP.git}}. Additional analyses related to part shareability and qualitative results are given in the Supplementary Material. 

\subsection{Analysis of Generative and Descriptive Properties}
We analyze the relationship between the number of classes, views, objects, and vocabulary size, average MDL values and test inference time in three different setups, respectively. Vocabulary size and test inference time analyses provide information about the part shareability and generative shape representation behavior of our algorithm (The inference time of CHOP is the average inference time on test images). We examine the variations of the average MDL values under different test sets. In order to get a more descriptive estimate of MDL values, we use 10 best parts constructed at each layer of CHOP. While a vocabulary layer may contain thousands of parts, most of the parts constructed with the lowest MDL scores belong to a single object in the model, and therefore exhibit no shareability. 

\textbf{Analyses with Different Number of Categories} In this section we use the first 30 categories of the MPEG-7 Core Experiment CE-Shape 1 dataset \cite{mpeg7}. We randomly select 5 images from each category to construct training sets.

\begin{figure}[t]
	\centering
	\subfloat[]{\includegraphics[scale=0.6]{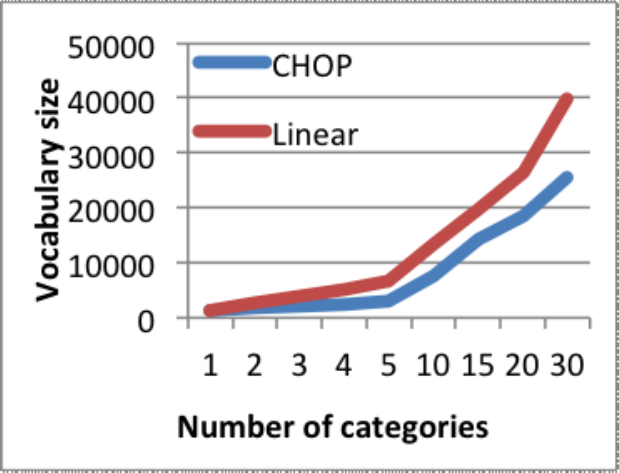}} \hspace{0.5mm}
	\subfloat[]{\includegraphics[scale=0.58]{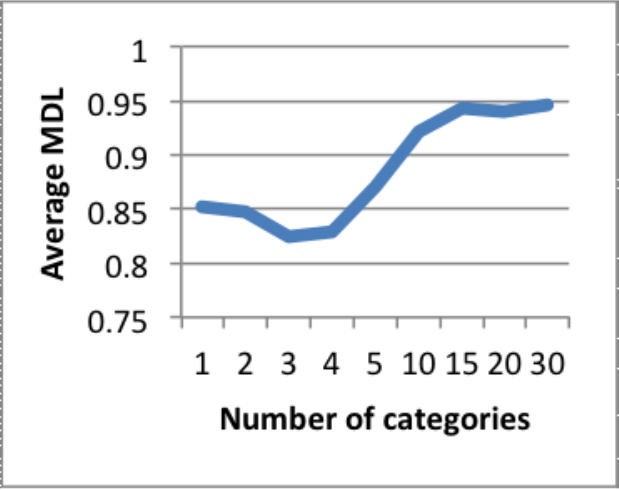}} \hspace{0.5mm}
	\subfloat[]{\includegraphics[scale=0.6]{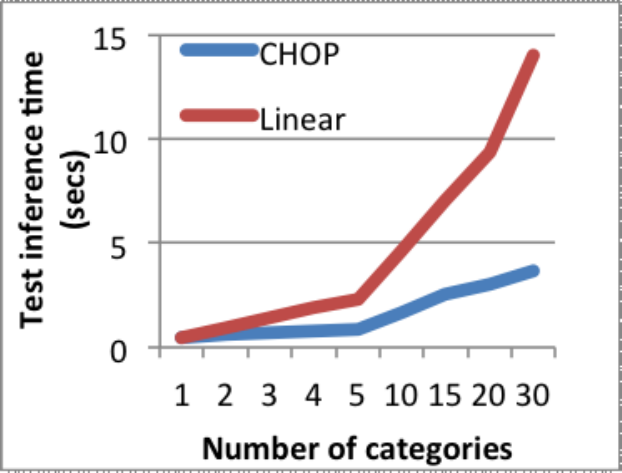}} \hspace{0.5mm}
	\caption{Analyses with different number of categories. (Best viewed in colour)}
	\label{fig:classvs}
\end{figure}

The vocabulary size grows sub-linearly as shown with the blue line in Fig. \ref{fig:classvs}.a. The higher part shareability observed in the first layers of CHOP is considered as the main contributing factor which affects the vocabulary size. We observe a sub-linear growth of the number of parts as the number of categories increases, which affects the test image inference time as shown in Fig. \ref{fig:classvs}.c. This is observed because the inference process requires searching every composition in the vocabulary within the graph representation of a test image. The  efficient indexing mechanism implemented in CHOP speeds up the testing time, and the average test time is calculated as ~0.5-3 seconds depending on the number of categories. Average MDL values tend to increase after a boost at around 3-4 categories (lower is better), and converge at 15 categories. The inter-class appearance differences allow for a limited amount of shareability between categories.

\subsection{Analyses with Different Number of Objects}
\begin{figure}[t]
	\centering
	\subfloat[]{\includegraphics[scale=0.6]{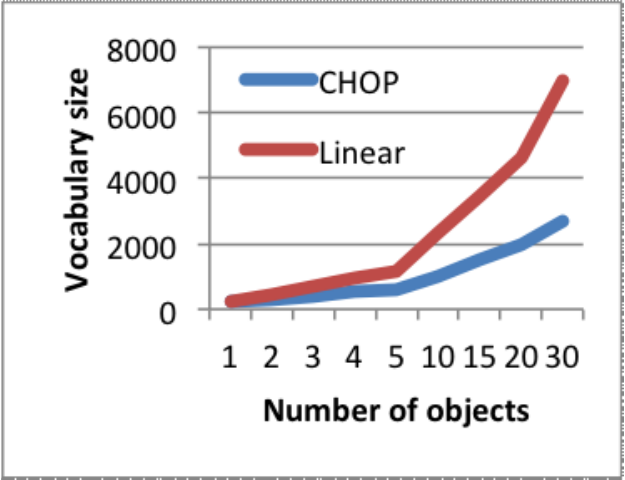}} \hspace{0.6mm}
	\subfloat[]{\includegraphics[scale=0.6]{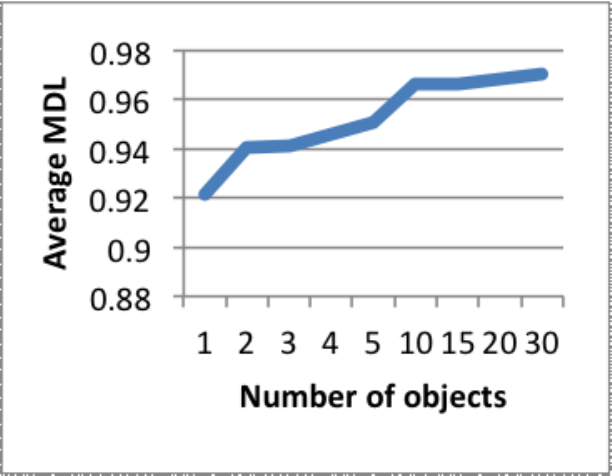}} \hspace{0.6mm}
	\subfloat[]{\includegraphics[scale=0.6]{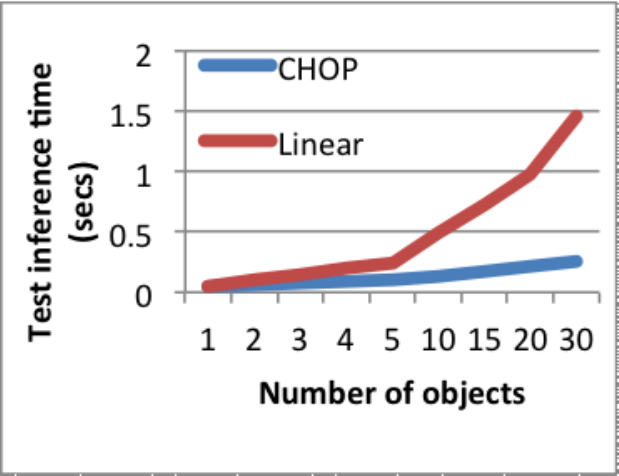}} \hspace{0.6mm}
	\caption{Analyses with different number of objects. (Best viewed in colour)}
	\label{fig:objectvs}
\end{figure}

In order to analyze the effect of increasing number of images to the proposed performance measures, we use 30 samples belonging to the "Apple Logos" class in ETHZ Shape Classes dataset \cite{ethshape} for training. Compared to the results obtained in the previous section, we observe that average MDL values increase gradually as the number of objects increase in Fig. \ref{fig:objectvs}.b. Additionally, the growth rate of the vocabulary size observed in Fig. \ref{fig:objectvs}.a is less than the one depicted in Fig. \ref{fig:classvs}.a.

\subsection{Analyses with Different Number of Views}
\begin{figure}[t]
	\centering
	\subfloat[]{\includegraphics[scale=0.6]{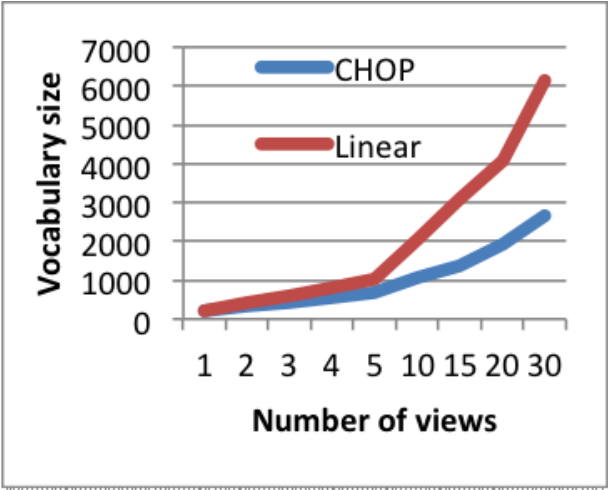}} \hspace{0.5mm}
	\subfloat[]{\includegraphics[scale=0.6]{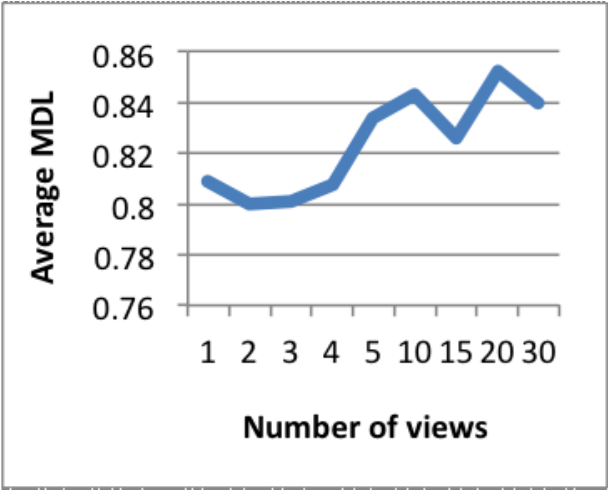}} \hspace{0.5mm}
	\subfloat[]{\includegraphics[scale=0.6]{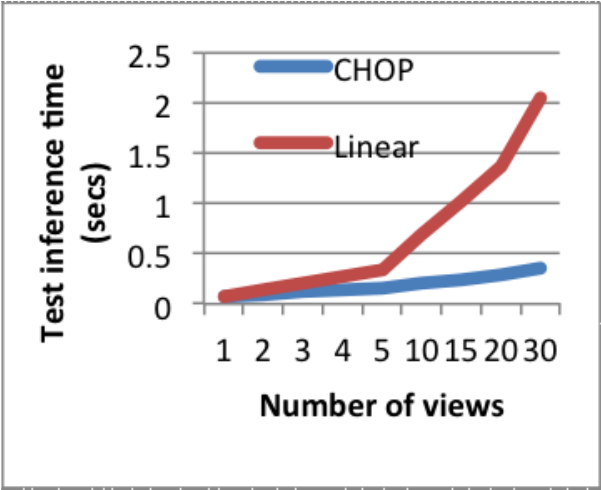}} \hspace{0.5mm}
	\caption{Analyses with different number of views. (Best viewed in colour)}
	\label{fig:viewvs}
\end{figure}

In the third set of experiments, we use a subset of Washington image dataset \cite{aloi} consisting of images captured at different views of the same object. Multiple view images of a cup are used as the training data. Due to the fairly symmetrical nature of a cup except for its textures and handle, the shareability of the parts in the vocabulary remains consistent as the training image set grows. Interestingly, we observe a local maximum at around 15 views in Fig. \ref{fig:viewvs}.b. Depending on the inhibition and part selection (SUBDUE) parameters, less frequently observed yet valuable parts may be discarded by the algorithm in mid-layers. 

\subsection{Shape Retrieval Experiments}


Following the results of \cite{raviv,Siddiqi,Siddiqi2}, we employ eigenvalues of adjacency matrices of edge weighted graphs computed using object graphs of shapes as shape descriptors. For this purpose, we first define edge weights $e_{ab} \in E_l$ of an edge weighted graph $W_l=(V_l,E_l) $ of an object graph $\mathbb{G}_l=(\mathbb{V}_l,\mathbb{E}_l)$ as 
\begin{equation}
e_{ab}= 
\begin{cases}
\pi_k, & \quad {\rm if} \quad { \rm R_a^l \; is \; connected \; to \; R_b^l}, \quad \forall{R_a^l,R_b^l} \in \mathbb{V}_l \ \\
0, & \quad otherwise
\end{cases} ,
\label{eq:H_edges}
\end{equation}
where  $\pi_k$ is the cluster index which minimizes the conditional entropy  \eqref{eq:cond_entropy} in \eqref{eq:cond_entropy_min}. Then, we compute the weighted adjacency matrix of $W_l$ and use the eigenvalues as shape descriptors. We compute the distance between two shapes as the Euclidean distance between their shape descriptors. 

In the first set of experiments, we compare the retrieval performances of CHOP and the state-of-the-art shape  classification algorithms which use inner-distance (ID) measures to compute shape descriptors which are robust to articulation \cite{tools-40}. The experiments are performed on Tools-40 dataset \cite{tools-40} which contains 40 images captured using 8 different objects each of which provides 5 articulated shapes. Given each query image, the four most similar matches are chosen from the other images in the dataset for the evaluation of the recognition results \cite{tools-40}. The results are summarized as the number of first, second, third and fourth most similar matches that come from the correct object in Table \ref{tab:tools_40}. We observe that CHOP provides better performance than the shape-based descriptors and retrieval algorithms SC+DP and MDS+SC+DP \cite{tools-40}. However, IDSC+DP \cite{tools-40}, which integrates texture information with the shape information, provides better performance for Top 1 retrieval results, and CHOP performs better than IDSC+DP for Top 4 retrieval results. The reason of this observation is that texture of shape structures provides discriminative information about shape categories. Therefore, the objects which have the most similar textures are closer to each other than the other objects as observed in Top 1 retrieval results. On the other hand, texture information may dominate the shape information and may lead to overfitting as observed in Top 4 retrieval results (see Table \ref{tab:tools_40}).

\begin{table}[t]
  \centering
  \caption{Comparison of shape retrieval performances (\%) on Tools-40 dataset.}
	\begin{tabular}{|c|c|c|c|c|}
    \hline
	\textbf{Algorithms} & Top 1 & Top 2 & Top 3 & Top 4 \\ 
	\hline
    SC+DP \cite{tools-40} & 20/40 & 10/40 & 11/40 & 5/40 \\
    MDS+SC+DP \cite{tools-40} & 36/40 & 26/40 & 17/40 & 15/40 \\
    IDSC+DP \cite{tools-40} & 40/40 & 34/40 & 35/40 & 27/40 \\
    CHOP &  37/40 &  35/40 &  35/40 &  29/40 \\
    \hline
    \end{tabular}%
  \label{tab:tools_40}%
\end{table}%

In the second set of experiments, we use Myth and Tools-35 datasets in order to analyze the performance of the shape retrieval algorithms \cite{prl} and CHOP, considering part shareability and category-wise articulation. In the Myth dataset, there are three categories, namely Centaur, Horse and Man, and 5 different images belonging to 5 different objects in each category. Shapes observed in images differ by articulation and additional parts, e.g. the shapes of objects belonging to Centaur and Man categories share the upper part of the man body, and the shapes of objects belonging to Centaur and Horse categories share the lower part of the horse body. In the Tools-35 dataset, there are 35 shapes belonging to 4 categories which are split as 10 scissors, 15 pliers, 5 pincers, 5 knives. Each object belonging to a category differs by an articulation. Performance values are calculated using a Bullseye test as suggested in \cite{prl} to compare the performances of CHOP and other shape retrieval algorithms Contour-ID \cite{prl} and Contour-HF \cite{prl}. In the Bullseye test, five most similar candidates for each query image are considered \cite{prl}. Experimental results given in Table \ref{tab:myth_35} show that CHOP outperforms  Contour-ID and Contour-HF \cite{prl} which employ distributions of descriptor values calculated at shape contours as shape features that are invariant to articulations and deformations in local part structures. However, part shareability and articulation properties of shapes may provide discriminative information about shape structures, especially on the images in the Myth dataset. 

\begin{table}[t]
  \centering
  \caption{Comparison of shape retrieval performances (\%) on Myth and Tools-35.}
	\begin{tabular}{|c|c|c|c|}
    \hline
	\textbf{Datasets} & Contour-ID \cite{prl} & Contour-HF \cite{prl} & CHOP \\ 
	\hline
    Tools-35 & 84.57 & 84.57 & 87.86 \\
    \hline
    Myth     & 77.33 & 90.67 & 93.33 \\
    \hline
    \end{tabular}%
  \label{tab:myth_35}%
\end{table}%


\section{Conclusion}
\label{sec:conclusion}
We have proposed a graph theoretic approach for object shape representation in a hierarchical compositional architecture called Compositional Hierarchy of Parts (CHOP). Two information theoretic algorithms are used for learning a vocabulary of compositional parts employing a hybrid generative-descriptive model. First, statistical relationships between parts are learned using the MCEC algorithm. Then, part selection problem is defined as a frequent subgraph discovery problem, and solved using an MDL principle. Part compositions are inferred considering both learned statistical relationships between parts and their description lengths at each layer of CHOP.

The proposed approach and algorithms are examined using six benchmark shape datasets consisting of different images of an object captured at different viewpoints, and images of objects belonging to different categories. The results show that CHOP can  use part shareability property in the construction of \textit{compact} vocabularies and inference trees efficiently. For instance, we observe that the running time of CHOP to perform inference on test images is approximately 0.5-3 seconds for an image. Additionally, we can construct compositional shape representations which provide part realizations that completely cover the shapes on the images. Finally, we compared shape retrieval performances of CHOP and the state-of-the-art retrieval algorithms on three benchmark datasets. The results show that CHOP outperforms the evaluated algorithms using part shareability and fast inference of descriptive part compositions.

In the future work, we will employ discriminative learning for pose estimation and categorization of shapes. In addition, online and incremental learning will be implemented considering the results obtained from the analyses on part shareability performed in this work. 

\section*{Acknowledgement}
This work was supported in part by the European Commission project PaCMan EU FP7-ICT, 600918. The authors would also like to thank Sebastian Zurek for helpful discussions.



%

\newpage
\bibliographystyle{IEEEtranS}
\bibliography{IEEEexample}

\end{document}


\pagestyle{headings}
\mainmatter

\def\ECCV14SubNumber{1558}  

\title{Supplementary Material \\
for \\
A Graph Theoretic Approach for Object Shape Representation in Compositional Hierarchies using a Hybrid Generative-Descriptive Model}

\titlerunning{Supplementary Material for ECCV-14 submission ID \ECCV14SubNumber}

\authorrunning{Supplementary Material for ECCV-14 submission ID \ECCV14SubNumber}

\author{Umit Rusen Aktas\thanks{The first and second author contributed equally.}, Mete Ozay\samethanks[1], Ale{\v s} Leonardis and Jeremy L. Wyatt
} 
\institute{School of Computer Science, The University of Birmingham, Edgbaston, Birmingham, B15 2TT, United Kingdom.
\\ Emails: \{ u.aktas, m.ozay, a.Leonardis, j.l.wyatt \} @cs.bham.ac.uk}


\maketitle



%

\section{Introduction}

In this supplementary material, we provide images of parts, compositions and realizations that are constructed and detected using Compositional Hierarchy of Parts (CHOP) algorithm which is introduced in the ECCV 2014 paper titled ``A Graph Theoretic Approach for Object Shape Representation in Compositional Hierarchies using a Hybrid Generative-Descriptive Model". 

Additionally, results which are obtained employing the CHOP on the images belonging to other datasets, and other images belonging to the datasets used in the ECCV 2014 paper, are given in addition to the ones that are provided in the paper.

\section{Experiments}
\label{sec:exp}

We examine our proposed approach on three benchmark object shape datasets, which are namely the Amsterdam Library of Object Images (ALOI) \cite{aloi}, the Tools and the Myth \cite{tools}. In the experiments \footnote{a Matlab implementation of CHOP is available on the webpage \url{https://github.com/rusen/CHOP.git}.}, we used $\Theta=6$ number of different orientations of Gabor features with the same Gabor kernel parameters implemented in \cite{fidler_cvpr07}. We used subsampling ratio as $\sigma=0.5$. In Section \ref{sec:shr}, we provide the results regarding shareability of parts that are constructed in the experiments presented in Section 4 of the main text of the ECCV 2014 paper. Section \ref{sec:mview} includes experimental results on vocabulary learning with images of objects that are captured from multiple-views.

\subsection{Results on Part Shareability}
\label{sec:shr}

In this section, we analyze the effect of increasing number of objects, views, and categories on the degree of shareability of parts in the vocabulary. The shareability of a part in a learned vocabulary is measured as the average number of objects, views and categories that the part is shared across multiple objects belonging to a category, multiple views of an object belonging to a category, and multiple objects belonging to multiple categories, respectively. The shareability of a vocabulary is defined as the average shareability of its parts. In order to consider the shareability of the \textit{most} descriptive parts in the analyses, we select 10 parts which have the best (lowest) MDL scores at each layer of the hierarchical vocabulary. Parts computed at the first layer $l=1$ are not considered in shareability analyses, because Gabor filter responses calculated at the first layer are shared across almost all images in all of the experiments.

\begin{figure}[!htbp]
	\centering
	\subfloat[Shareability vs Number of Objects]{\includegraphics[scale=0.57]{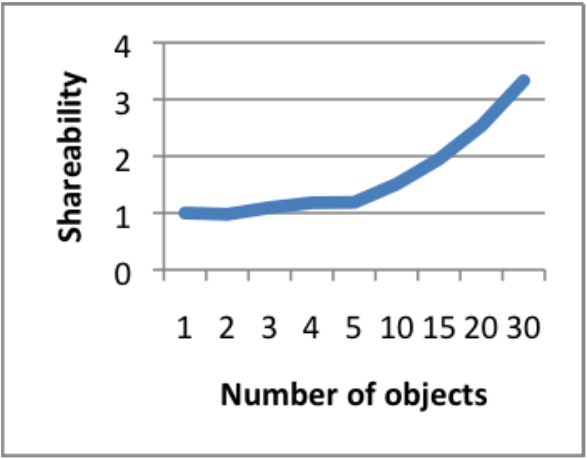}} \hspace{0.5cm}
	\subfloat[Shareability vs Number of Views]{\includegraphics[scale=0.6]{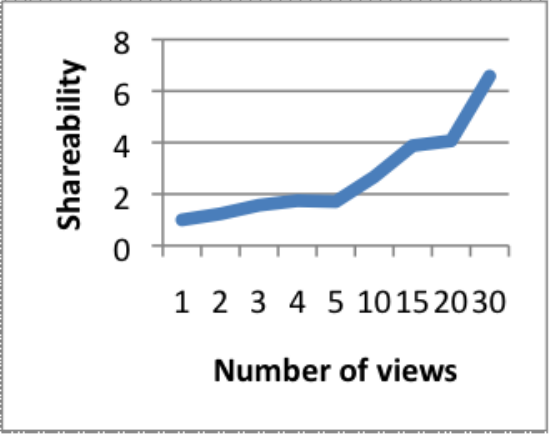}} \hspace{0.5cm}
	\subfloat[Shareability vs Number of Categories]{\includegraphics[scale=0.57]{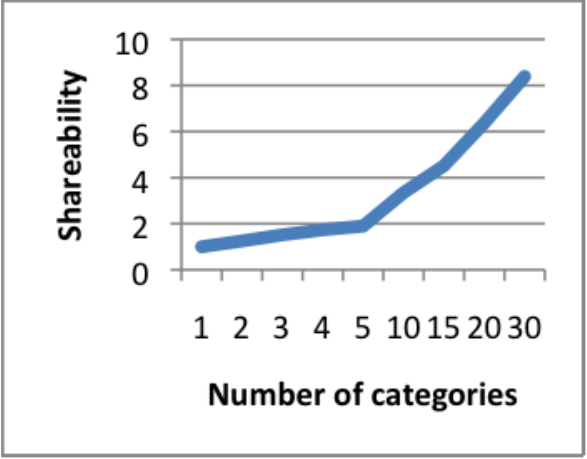}} \hspace{0.5cm}
	\caption{Shareability analysis of the three sets of experiments in ECCV 2014 paper.}
	\label{fig:shr}
\end{figure}

The relationship between the number of objects belonging to a single class and the shareability property of the shape model is illustrated in Fig. \ref{fig:shr}.a. The results show that the shareability of parts increases as the number of objects increases. The images used in this experiment are selected from the ``Apple Logos" class in the ETHZ Shape Classes dataset \cite{ethshape}. The instances exhibit high variability in scale and rotation, therefore reducing part shareability. Similarly, in Fig. \ref{fig:shr}.b, we analyze the shareability of parts as the number of views of the same object increases. In this set of experiments, we selected a cup from Washington image dataset \cite{aloi}, which is used in Section 4.3 of the ECCV paper. Since similar parts are constructed using the images of objects captured at different views, we obtain higher shareability scores when compared to the single-category case in Fig. \ref{fig:shr}.a. Finally, Fig. \ref{fig:shr}.c illustrates the effect of training a vocabulary with objects belonging to different number of categories on the shareability of parts. 

\newpage

\subsection{Experiments on Multiple-View Images}
\label{sec:mview}

Amsterdam Library of Object Images (ALOI) \cite{aloiReal} dataset consists of multiple view images of objects belonging to 1000 categories. In the experiments presented in this section, we used 14 images captured from the viewpoints labelled $\{ 25^o, 50^o, 75^o, \ldots, 350^o \}$  as test images, and 14 images captured from the viewpoints labelled $\{ 30^o, 55^o, 80^o, \ldots, 355^o \}$  as training images, for each object. 

In the first set of experiments, we analyzed the part shareability and computational complexity of the algorithms across multiple view images of a cup and a duck. For each layer $l=1,2,3,4,5$, part realizations and object graphs detected on multiple view cup images and duck images are shown in Table \ref{tab:aloi_cups} and Table \ref{tab:aloi_duck}, respectively. In the images, each part with a different part realization id is depicted by a different color. For instance, for an image of a cup captured from a viewpoint labelled as $75^o$, there are 6 different types of parts with 78 different part realizations at the first layer $l=1$ (see second column of Table \ref{tab:aloi_cups}). However, we observe 5 different types of part compositions at the fifth layer $l=5$ of the hierarchy. In the results, each node of an object graph, which is visualized by red points and lines, represents the position of the center of a part. 

In the analyses of graph structures, we observe that the locality of topological structures of object graphs decreases through the higher layers representing object shapes with higher abstraction. For instance, part realizations of the parts represented with Gabor features at the first layer are connected to each other in a spatial neighbourhood in the results shown at $l=1$ and $l=2$ in Table \ref{tab:aloi_cups} and Table \ref{tab:aloi_duck}. However, the connectivity of part realizations is determined using statistical and descriptive relationships between parts at the higher layers; horizontally oriented part realizations detected at the top and bottom of a cup and a duck are connected to each other, and vertically oriented part realizations detected at the right and left of the cup and duck are connected to each other for $l \geq 3$ in Table \ref{tab:aloi_cups} and Table \ref{tab:aloi_duck}. 

\newpage


\begin{table}[h!]

  \caption{Results on multiple view cup images obtained from ALOI Dataset} 
  \label{tab:aloi_cups}
  
  \begin{tabular}
      {|l|l|l|l|l|l|l|} \hline \textbf{Rotation Degree} & $25^o$ & $75^o$ & $150^o$ &$225^o$ & $300^o$ & $350^o$ \\
   
      \hline \textbf{Original Image} &  
      \includegraphics[scale=0.2]{images/cups/125_r25-eps-converted-to.pdf} & \includegraphics[scale=0.2]{images/cups/125_r75-eps-converted-to.pdf}& \includegraphics[scale=0.2]{images/cups/125_r150-eps-converted-to.pdf} & \includegraphics[scale=0.2]{images/cups/125_r225-eps-converted-to.pdf} & \includegraphics[scale=0.2]{images/cups/125_r300-eps-converted-to.pdf}  & \includegraphics[scale=0.2]{images/cups/125_r350-eps-converted-to.pdf}    \\
      
      \hline \textbf{Layer $l=1$} &  
      \includegraphics[scale=0.2]{images/cups/125_r25_level1-eps-converted-to.pdf} & \includegraphics[scale=0.2]{images/cups/125_r75_level1-eps-converted-to.pdf} & \includegraphics[scale=0.2]{images/cups/125_r150_level1-eps-converted-to.pdf} & \includegraphics[scale=0.2]{images/cups/125_r225_level1-eps-converted-to.pdf} & \includegraphics[scale=0.2]{images/cups/125_r300_level1-eps-converted-to.pdf}  & \includegraphics[scale=0.2]{images/cups/125_r350_level1-eps-converted-to.pdf}    \\
      
      \hline \textbf{Object Graph at $l=1$} &  
      \includegraphics[scale=0.2]{images/cups/125_r25_level1edges-eps-converted-to.pdf} & \includegraphics[scale=0.2]{images/cups/125_r75_level1edges-eps-converted-to.pdf} & \includegraphics[scale=0.2]{images/cups/125_r150_level1edges-eps-converted-to.pdf} & \includegraphics[scale=0.2]{images/cups/125_r225_level1edges-eps-converted-to.pdf} & \includegraphics[scale=0.2]{images/cups/125_r300_level1edges-eps-converted-to.pdf}  & \includegraphics[scale=0.2]{images/cups/125_r350_level1edges-eps-converted-to.pdf}    \\
      
      \hline \textbf{Layer $l=2$} &  
      \includegraphics[scale=0.2]{images/cups/125_r25_level2-eps-converted-to.pdf} & \includegraphics[scale=0.2]{images/cups/125_r75_level2-eps-converted-to.pdf} & \includegraphics[scale=0.2]{images/cups/125_r150_level2-eps-converted-to.pdf} & \includegraphics[scale=0.2]{images/cups/125_r225_level2-eps-converted-to.pdf} & \includegraphics[scale=0.2]{images/cups/125_r300_level2-eps-converted-to.pdf}  & \includegraphics[scale=0.2]{images/cups/125_r350_level2-eps-converted-to.pdf}    \\
      
      \hline \textbf{Object Graph at $l=2$} &  
      \includegraphics[scale=0.2]{images/cups/125_r25_level2edges-eps-converted-to.pdf} & \includegraphics[scale=0.2]{images/cups/125_r75_level2edges-eps-converted-to.pdf} & \includegraphics[scale=0.2]{images/cups/125_r150_level2edges-eps-converted-to.pdf} & \includegraphics[scale=0.2]{images/cups/125_r225_level2edges-eps-converted-to.pdf} & \includegraphics[scale=0.2]{images/cups/125_r300_level2edges-eps-converted-to.pdf}  & \includegraphics[scale=0.2]{images/cups/125_r350_level2edges-eps-converted-to.pdf}    \\
      
      \hline \textbf{Layer $l=3$} &  
      \includegraphics[scale=0.2]{images/cups/125_r25_level3-eps-converted-to.pdf} & \includegraphics[scale=0.2]{images/cups/125_r75_level3-eps-converted-to.pdf} & \includegraphics[scale=0.2]{images/cups/125_r150_level3-eps-converted-to.pdf} & \includegraphics[scale=0.2]{images/cups/125_r225_level3-eps-converted-to.pdf} & \includegraphics[scale=0.2]{images/cups/125_r300_level3-eps-converted-to.pdf}  & \includegraphics[scale=0.2]{images/cups/125_r350_level3-eps-converted-to.pdf}    \\
      
      \hline \textbf{Object Graph at $l=3$} & 
      \includegraphics[scale=0.2]{images/cups/125_r25_level3edges-eps-converted-to.pdf} & \includegraphics[scale=0.2]{images/cups/125_r75_level3edges-eps-converted-to.pdf} & \includegraphics[scale=0.2]{images/cups/125_r150_level3edges-eps-converted-to.pdf} & \includegraphics[scale=0.2]{images/cups/125_r225_level3edges-eps-converted-to.pdf} & \includegraphics[scale=0.2]{images/cups/125_r300_level3edges-eps-converted-to.pdf}  & \includegraphics[scale=0.2]{images/cups/125_r350_level3edges-eps-converted-to.pdf}    \\
      
      \hline \textbf{Layer $l=4$} &  
      \includegraphics[scale=0.2]{images/cups/125_r25_level4-eps-converted-to.pdf} & \includegraphics[scale=0.2]{images/cups/125_r75_level4-eps-converted-to.pdf} & \includegraphics[scale=0.2]{images/cups/125_r150_level4-eps-converted-to.pdf} & \includegraphics[scale=0.2]{images/cups/125_r225_level4-eps-converted-to.pdf} & \includegraphics[scale=0.2]{images/cups/125_r300_level4-eps-converted-to.pdf}  & \includegraphics[scale=0.2]{images/cups/125_r350_level4-eps-converted-to.pdf}    \\
      
      \hline \textbf{Object Graph at $l=4$} &  
      \includegraphics[scale=0.2]{images/cups/125_r25_level4edges-eps-converted-to.pdf} & \includegraphics[scale=0.2]{images/cups/125_r75_level4edges-eps-converted-to.pdf} & \includegraphics[scale=0.2]{images/cups/125_r150_level4edges-eps-converted-to.pdf} & \includegraphics[scale=0.2]{images/cups/125_r225_level4edges-eps-converted-to.pdf} & \includegraphics[scale=0.2]{images/cups/125_r300_level4edges-eps-converted-to.pdf}  & \includegraphics[scale=0.2]{images/cups/125_r350_level4edges-eps-converted-to.pdf}    \\
      
      \hline \textbf{Layer $l=5$} &  
      \includegraphics[scale=0.2]{images/cups/125_r25_level5-eps-converted-to.pdf} & \includegraphics[scale=0.2]{images/cups/125_r75_level5-eps-converted-to.pdf} & \includegraphics[scale=0.2]{images/cups/125_r150_level5-eps-converted-to.pdf} & \includegraphics[scale=0.2]{images/cups/125_r225_level5-eps-converted-to.pdf} & \includegraphics[scale=0.2]{images/cups/125_r300_level5-eps-converted-to.pdf}  & \includegraphics[scale=0.2]{images/cups/125_r350_level5-eps-converted-to.pdf}    \\
      
      \hline \textbf{Object Graph at $l=5$} & 
      \includegraphics[scale=0.2]{images/cups/125_r25_level5edges-eps-converted-to.pdf} & \includegraphics[scale=0.2]{images/cups/125_r75_level5edges-eps-converted-to.pdf} & \includegraphics[scale=0.2]{images/cups/125_r150_level5edges-eps-converted-to.pdf} & \includegraphics[scale=0.2]{images/cups/125_r225_level5edges-eps-converted-to.pdf} & \includegraphics[scale=0.2]{images/cups/125_r300_level5edges-eps-converted-to.pdf}  & \includegraphics[scale=0.2]{images/cups/125_r350_level5edges-eps-converted-to.pdf}    \\
      
      \hline
  \end{tabular}
\end{table}

\begin{table}[ht!]
\caption{Results on multiple view duck images obtained from ALOI Dataset} \label{tab:aloi_duck}
  \begin{tabular}
      {|l|l|l|l|l|l|l|} \hline \textbf{Rotation Degree} & $25^o$ & $75^o$ & $150^o$ &$225^o$ & $300^o$ & $350^o$ \\
      \hline \textbf{Original Image} &  
      \includegraphics[scale=0.2]{images/duck/90_r25-eps-converted-to.pdf} & \includegraphics[scale=0.2]{images/duck/90_r75-eps-converted-to.pdf} & \includegraphics[scale=0.2]{images/duck/90_r150-eps-converted-to.pdf} & \includegraphics[scale=0.2]{images/duck/90_r225-eps-converted-to.pdf}& \includegraphics[scale=0.2]{images/duck/90_r300-eps-converted-to.pdf}  & \includegraphics[scale=0.2]{images/duck/90_r350-eps-converted-to.pdf}    \\
      
      \hline \textbf{Layer $l=1$} &  
      \includegraphics[scale=0.2]{images/duck/90_r25_level1-eps-converted-to.pdf} & \includegraphics[scale=0.2]{images/duck/90_r75_level1-eps-converted-to.pdf} & \includegraphics[scale=0.2]{images/duck/90_r150_level1-eps-converted-to.pdf} & \includegraphics[scale=0.2]{images/duck/90_r225_level1-eps-converted-to.pdf} & \includegraphics[scale=0.2]{images/duck/90_r300_level1-eps-converted-to.pdf}  & \includegraphics[scale=0.2]{images/duck/90_r350_level1-eps-converted-to.pdf}    \\
      
      \hline \textbf{Object Graph at $l=1$} &  
      \includegraphics[scale=0.2]{images/duck/90_r25_level1edges-eps-converted-to.pdf} & \includegraphics[scale=0.2]{images/duck/90_r75_level1edges-eps-converted-to.pdf} & \includegraphics[scale=0.2]{images/duck/90_r150_level1edges-eps-converted-to.pdf} & \includegraphics[scale=0.2]{images/duck/90_r225_level1edges-eps-converted-to.pdf} & \includegraphics[scale=0.2]{images/duck/90_r300_level1edges-eps-converted-to.pdf}  & \includegraphics[scale=0.2]{images/duck/90_r350_level1edges-eps-converted-to.pdf}    \\
      
      \hline \textbf{Layer $l=2$} &  
      \includegraphics[scale=0.2]{images/duck/90_r25_level2-eps-converted-to.pdf} & \includegraphics[scale=0.2]{images/duck/90_r75_level2-eps-converted-to.pdf} & \includegraphics[scale=0.2]{images/duck/90_r150_level2-eps-converted-to.pdf} & \includegraphics[scale=0.2]{images/duck/90_r225_level2-eps-converted-to.pdf} & \includegraphics[scale=0.2]{images/duck/90_r300_level2-eps-converted-to.pdf}  & \includegraphics[scale=0.2]{images/duck/90_r350_level2-eps-converted-to.pdf}    \\
      
      \hline \textbf{Object Graph at $l=2$} &  
      \includegraphics[scale=0.2]{images/duck/90_r25_level2edges-eps-converted-to.pdf} & \includegraphics[scale=0.2]{images/duck/90_r75_level2edges-eps-converted-to.pdf} & \includegraphics[scale=0.2]{images/duck/90_r150_level2edges-eps-converted-to.pdf} & \includegraphics[scale=0.2]{images/duck/90_r225_level2edges-eps-converted-to.pdf} & \includegraphics[scale=0.2]{images/duck/90_r300_level2edges-eps-converted-to.pdf}  & \includegraphics[scale=0.2]{images/duck/90_r350_level2edges-eps-converted-to.pdf}    \\
      
      \hline \textbf{Layer $l=3$} &  
      \includegraphics[scale=0.2]{images/duck/90_r25_level3-eps-converted-to.pdf} & \includegraphics[scale=0.2]{images/duck/90_r75_level3-eps-converted-to.pdf} & \includegraphics[scale=0.2]{images/duck/90_r150_level3-eps-converted-to.pdf} & \includegraphics[scale=0.2]{images/duck/90_r225_level3-eps-converted-to.pdf} & \includegraphics[scale=0.2]{images/duck/90_r300_level3-eps-converted-to.pdf}  & \includegraphics[scale=0.2]{images/duck/90_r350_level3-eps-converted-to.pdf}    \\
      
      \hline \textbf{Object Graph at $l=3$} & 
      \includegraphics[scale=0.2]{images/duck/90_r25_level3edges-eps-converted-to.pdf} & \includegraphics[scale=0.2]{images/duck/90_r75_level3edges-eps-converted-to.pdf} & \includegraphics[scale=0.2]{images/duck/90_r150_level3edges-eps-converted-to.pdf} & \includegraphics[scale=0.2]{images/duck/90_r225_level3edges-eps-converted-to.pdf} & \includegraphics[scale=0.2]{images/duck/90_r300_level3edges-eps-converted-to.pdf}  & \includegraphics[scale=0.2]{images/duck/90_r350_level3edges-eps-converted-to.pdf}    \\
      
      \hline \textbf{Layer $l=4$} &  
      \includegraphics[scale=0.2]{images/duck/90_r25_level4-eps-converted-to.pdf} & \includegraphics[scale=0.2]{images/duck/90_r75_level4-eps-converted-to.pdf} & \includegraphics[scale=0.2]{images/duck/90_r150_level4-eps-converted-to.pdf} & \includegraphics[scale=0.2]{images/duck/90_r225_level4-eps-converted-to.pdf} & \includegraphics[scale=0.2]{images/duck/90_r300_level4-eps-converted-to.pdf}  & \includegraphics[scale=0.2]{images/duck/90_r350_level4-eps-converted-to.pdf}    \\
      
      \hline \textbf{Object Graph at $l=4$} &  
      \includegraphics[scale=0.2]{images/duck/90_r25_level4edges-eps-converted-to.pdf} & \includegraphics[scale=0.2]{images/duck/90_r75_level4edges-eps-converted-to.pdf} & \includegraphics[scale=0.2]{images/duck/90_r150_level4edges-eps-converted-to.pdf} & \includegraphics[scale=0.2]{images/duck/90_r225_level4edges-eps-converted-to.pdf} & \includegraphics[scale=0.2]{images/duck/90_r300_level4edges-eps-converted-to.pdf}  & \includegraphics[scale=0.2]{images/duck/90_r350_level4edges-eps-converted-to.pdf}    \\
      
      \hline \textbf{Layer $l=5$} &  
      \includegraphics[scale=0.2]{images/duck/90_r25_level5-eps-converted-to.pdf} & \includegraphics[scale=0.2]{images/duck/90_r75_level5-eps-converted-to.pdf} & \includegraphics[scale=0.2]{images/duck/90_r150_level5-eps-converted-to.pdf} & \includegraphics[scale=0.2]{images/duck/90_r225_level5-eps-converted-to.pdf} & \includegraphics[scale=0.2]{images/duck/90_r300_level5-eps-converted-to.pdf}  & \includegraphics[scale=0.2]{images/duck/90_r350_level5-eps-converted-to.pdf}    \\
      
      \hline \textbf{Object Graph at $l=5$} & 
      \includegraphics[scale=0.2]{images/duck/90_r25_level5edges-eps-converted-to.pdf} & \includegraphics[scale=0.2]{images/duck/90_r75_level5edges-eps-converted-to.pdf} & \includegraphics[scale=0.2]{images/duck/90_r150_level5edges-eps-converted-to.pdf} & \includegraphics[scale=0.2]{images/duck/90_r225_level5edges-eps-converted-to.pdf} & \includegraphics[scale=0.2]{images/duck/90_r300_level5edges-eps-converted-to.pdf}  & \includegraphics[scale=0.2]{images/duck/90_r350_level5edges-eps-converted-to.pdf}    \\
      
      \hline \textbf{Layer $l=6$} &  
      \includegraphics[scale=0.2]{images/duck/90_r25_level6-eps-converted-to.pdf} & \includegraphics[scale=0.2]{images/duck/90_r75_level6-eps-converted-to.pdf} & \includegraphics[scale=0.2]{images/duck/90_r150_level6-eps-converted-to.pdf} & \includegraphics[scale=0.2]{images/duck/90_r225_level6-eps-converted-to.pdf} & \includegraphics[scale=0.2]{images/duck/90_r300_level6-eps-converted-to.pdf}  & \includegraphics[scale=0.2]{images/duck/90_r350_level6-eps-converted-to.pdf}    \\
      
      \hline \textbf{Object Graph at $l=6$} & 
      \includegraphics[scale=0.2]{images/duck/90_r25_level6edges-eps-converted-to.pdf} & \includegraphics[scale=0.2]{images/duck/90_r75_level6edges-eps-converted-to.pdf} & \includegraphics[scale=0.2]{images/duck/90_r150_level6edges-eps-converted-to.pdf} & \includegraphics[scale=0.2]{images/duck/90_r225_level6edges-eps-converted-to.pdf} & \includegraphics[scale=0.2]{images/duck/90_r300_level6edges-eps-converted-to.pdf}  & \includegraphics[scale=0.2]{images/duck/90_r350_level6edges-eps-converted-to.pdf}   \\      
      \hline
  \end{tabular} 
\end{table}

\newpage
\clearpage


\subsection{Experiments on Partial Shape Similarity}

Employing part shape similarity for learning compositions of parts is an important requirement for hierarchical compositional architectures \cite{lhop_book}. In this section, we examine this property of the proposed CHOP algorithm in an articulated shape dataset called the Myth dataset \cite{tools}.

In the Myth dataset, there are three categories, namely \textit{Centaur}, \textit{Horse} and \textit{Man}. There are 5 different images belonging to 5 different objects in each category. Shapes observed in images differ by additional parts, e.g. the shapes of the objects belonging to Centaur and Man categories share the upper part of a man's body, and the shapes of the objects belonging to Centaur and Horse categories share the lower part of a horse's body. In the experiments, four samples belonging to each category is used for training and the other three images are used for testing. The results of four experiments are shown in Table \ref{tab:myth_centaur}, \ref{tab:myth_horse}, \ref{tab:myth_man} for Centaur, Horse and Man categories, respectively. The results are shown for the last two layers that are achieved in the construction of object graphs for each shape. In the tables, the right column labeled $l+1$ represents the top layer, and the left column labeled $l$ represents the previous column. For instance, the left column of Centaur-1 shape depicts part realizations and object graphs detected at the layer $l = 7$, and the right column depicts part realizations and object graphs detected at the layer $l+1 = 8$ of the hierarchy in Table  \ref{tab:myth_centaur}. Note that top layers of inference trees at which part realizations and object graphs are detected can be different for different shapes and images, since a hierarchical vocabulary and inference trees are dynamically constructed in the CHOP.

In the experiments, we first observe that the depths of inference trees of the objects belonging to the same category are closer to each other than those of the objects belonging to different categories. For instance, the depth of inference trees for three Centaur shapes are $8$ and that of one Centaur shape is $7$. Meanwhile, the depth of inference trees of three Man shapes are $6$ and that of one Man shape is $7$.

Moreover, we observe that the shared parts are correctly detected in part realizations and successfully employed in the construction of compositions. For instance, legs of horses which are shared among Centaur and Horse categories are represented as single compositions in the vocabularies and detected as realizations with unique ids at the top layer of the inference trees. However, back parts of horses are depicted with different shapes, therefore these parts are not shared across categories. Consequently, the unshared parts are not detected in the inference trees and are not of great significance in the construction of part vocabularies. Similarly, the articulated right arms of man shapes which are shared across five shapes belonging to Man and Centaur categories are detected in the inference trees.

\begin{table}[ht!]
\caption{Results on images belonging to Centaur category obtained from Myth Dataset} \label{tab:myth_centaur}
  \begin{tabular}{|>{\centering\arraybackslash}m{1.8cm}|>{\centering\arraybackslash}m{2.5cm}|>{\centering\arraybackslash}m{2.5cm}|>{\centering\arraybackslash}m{2.5cm}|>{\centering\arraybackslash}m{2.5cm}|} 
  \hline
  \textbf{Object Name}, \textit{Layer ID} & \multicolumn{2}{|c|}{\centering $l$} & \multicolumn{2}{|c|}{\centering $l+1$} \\
      \hline & Part Realizations $R ^l$ & Object Graph $\mathbb{G}_l$ & Part Realizations $R ^{l+1}$ & Object Graph $\mathbb{G}_{l+1}$ \\
      \hline \centering \textbf{Centaur-1, $l=7$} &  
      \includegraphics[scale=0.2]{images/myth/centaur1_level7-eps-converted-to.pdf} & \includegraphics[scale=0.2]{images/myth/centaur1_level7edges-eps-converted-to.pdf} & \includegraphics[scale=0.2]{images/myth/centaur1_level8-eps-converted-to.pdf} & \includegraphics[scale=0.2]{images/myth/centaur1_level8edges-eps-converted-to.pdf}   \\
      
      \hline \textbf{Centaur-2, $l=7$} &  
      \includegraphics[scale=0.2]{images/myth/centaur2_level7-eps-converted-to.pdf} & \includegraphics[scale=0.2]{images/myth/centaur2_level7edges-eps-converted-to.pdf} & \includegraphics[scale=0.2]{images/myth/centaur2_level8-eps-converted-to.pdf} & \includegraphics[scale=0.2]{images/myth/centaur2_level8edges-eps-converted-to.pdf}   \\

	   \hline \centering \centering \textbf{Centaur-3, $l=7$} &  
      \includegraphics[scale=0.2]{images/myth/centaur3_level7-eps-converted-to.pdf} & \includegraphics[scale=0.2]{images/myth/centaur3_level7edges-eps-converted-to.pdf} & \includegraphics[scale=0.2]{images/myth/centaur3_level8-eps-converted-to.pdf} & \includegraphics[scale=0.2]{images/myth/centaur3_level8edges-eps-converted-to.pdf}   \\   
      
      \hline \centering \textbf{Centaur-4, $l=6$} &  
      \includegraphics[scale=0.2]{images/myth/centaur4_level6-eps-converted-to.pdf} & \includegraphics[scale=0.2]{images/myth/centaur4_level6edges-eps-converted-to.pdf} & \includegraphics[scale=0.2]{images/myth/centaur4_level7-eps-converted-to.pdf} & \includegraphics[scale=0.2]{images/myth/centaur4_level7edges-eps-converted-to.pdf}   \\
      \hline
  \end{tabular}
 \end{table}

\begin{table}[ht!]
\caption{Results on images belonging to Horse category obtained from Myth Dataset} \label{tab:myth_horse}
    \begin{tabular}{|>{\centering\arraybackslash}m{1.8cm}|>{\centering\arraybackslash}m{2.5cm}|>{\centering\arraybackslash}m{2.5cm}|>{\centering\arraybackslash}m{2.5cm}|>{\centering\arraybackslash}m{2.5cm}|}  
  \hline
  \textbf{Object Name}, \textit{Layer ID} & \multicolumn{2}{|c|}{$l$} & \multicolumn{2}{|c|}{$l+1$} \\
      \hline & Part Realizations $R ^l$ & Object Graph $\mathbb{G}_l$ & Part Realizations $R ^{l+1}$ & Object Graph $\mathbb{G}_{l+1}$ \\
      
      \hline \textbf{Horse-1, $l=7$} &  
      \includegraphics[scale=0.2]{images/myth/horse1_level7-eps-converted-to.pdf} & \includegraphics[scale=0.2]{images/myth/horse1_level7edges-eps-converted-to.pdf} & \includegraphics[scale=0.2]{images/myth/horse1_level8-eps-converted-to.pdf} & \includegraphics[scale=0.2]{images/myth/horse1_level8edges-eps-converted-to.pdf}   \\
      
      \hline \textbf{Horse-2, $l=7$} &  
      \includegraphics[scale=0.2]{images/myth/horse2_level7-eps-converted-to.pdf} & \includegraphics[scale=0.2]{images/myth/horse2_level7edges-eps-converted-to.pdf} & \includegraphics[scale=0.2]{images/myth/horse2_level8-eps-converted-to.pdf} & \includegraphics[scale=0.2]{images/myth/horse2_level8edges-eps-converted-to.pdf}   \\
      
      
      \hline \textbf{Horse-3, $l=6$} &  
      \includegraphics[scale=0.2]{images/myth/horse4_level6-eps-converted-to.pdf} & \includegraphics[scale=0.2]{images/myth/horse4_level6edges-eps-converted-to.pdf} & \includegraphics[scale=0.2]{images/myth/horse4_level7-eps-converted-to.pdf} & \includegraphics[scale=0.2]{images/myth/horse4_level7edges-eps-converted-to.pdf}   \\
      
      \hline \textbf{Horse-4, $l=6$} &  
      \includegraphics[scale=0.2]{images/myth/horse5_level6-eps-converted-to.pdf} & \includegraphics[scale=0.2]{images/myth/horse5_level6edges-eps-converted-to.pdf} & \includegraphics[scale=0.2]{images/myth/horse5_level7-eps-converted-to.pdf} & \includegraphics[scale=0.2]{images/myth/horse5_level7edges-eps-converted-to.pdf}   \\

      \hline
  \end{tabular}
  \end{table}

\begin{table}[ht!]
\caption{Results on images belonging to Man category obtained from Myth Dataset} \label{tab:myth_man}
    \begin{tabular}{|>{\centering\arraybackslash}m{1.8cm}|>{\centering\arraybackslash}m{2.5cm}|>{\centering\arraybackslash}m{2.5cm}|>{\centering\arraybackslash}m{2.5cm}|>{\centering\arraybackslash}m{2.5cm}|}  
  \hline
  \textbf{Object Name}, \textit{Layer ID} & \multicolumn{2}{|c|}{$l$} & \multicolumn{2}{|c|}{$l+1$} \\
      \hline & Part Realizations $R ^l$ & Object Graph $\mathbb{G}_l$ & Part Realizations $R ^{l+1}$ & Object Graph $\mathbb{G}_{l+1}$ \\
         
      \hline \textbf{Man-1, $l=5$} &  
      \includegraphics[scale=0.2]{images/myth/man1_level5-eps-converted-to.pdf} & \includegraphics[scale=0.2]{images/myth/man1_level5edges-eps-converted-to.pdf} & \includegraphics[scale=0.2]{images/myth/man1_level6-eps-converted-to.pdf} & \includegraphics[scale=0.2]{images/myth/man1_level6edges-eps-converted-to.pdf}   \\
      
      \hline \textbf{Man-2, $l=5$} &  
      \includegraphics[scale=0.2]{images/myth/man2_level5-eps-converted-to.pdf} & \includegraphics[scale=0.2]{images/myth/man2_level5edges-eps-converted-to.pdf} & \includegraphics[scale=0.2]{images/myth/man2_level6-eps-converted-to.pdf} & \includegraphics[scale=0.2]{images/myth/man2_level6edges-eps-converted-to.pdf}   \\
      
      \hline \textbf{Man-3, $l=5$} &  
      \includegraphics[scale=0.2]{images/myth/man4_level5-eps-converted-to.pdf} & \includegraphics[scale=0.2]{images/myth/man4_level5edges-eps-converted-to.pdf} & \includegraphics[scale=0.2]{images/myth/man4_level6-eps-converted-to.pdf} & \includegraphics[scale=0.2]{images/myth/man4_level6edges-eps-converted-to.pdf}   \\
      
      \hline \textbf{Man-4, $l=6$} &  
      \includegraphics[scale=0.2]{images/myth/man5_level6-eps-converted-to.pdf} & \includegraphics[scale=0.2]{images/myth/man5_level6edges-eps-converted-to.pdf} & \includegraphics[scale=0.2]{images/myth/man5_level7-eps-converted-to.pdf} & \includegraphics[scale=0.2]{images/myth/man5_level7edges-eps-converted-to.pdf}   \\
      
      \hline
  \end{tabular}
  
\end{table}

\newpage


\subsection{Experiments on Articulated Shape Images}

In the last set of experiments, we examined the proposed approach using the articulated Tools dataset \cite{tools}. The dataset consists of 35 shapes belonging to 4 categories. Images belonging to Scissors and Pliers categories are used in the experiments. In each experiment, we selected one object belonging to a category as a training object and the other object in the same category as a test object. An articulation is used to construct different shapes of objects. Experiments on Scissors and Pliers categories are shown in Table \ref{tab:tools_sci1} and \ref{tab:tools_sci2}, and Table \ref{tab:tools_pliers1} and \ref{tab:tools_pliers2}, respectively. For instance, the images belonging to Scissors-2 are used for training a vocabulary of a CHOP for detection of parts of shapes in images belonging to Scissors-1 in the experiments given in Table \ref{tab:tools_sci1}, and vice versa in Table \ref{tab:tools_sci2}.

In the results, junctions and closed curves observed at the shape boundaries are detected as part realizations, if they are shared among different articulated images. Moreover, these shape parts are represented as single part compositions at the top layers of inference trees by object graphs. For instance, circular shape handles of scissors  and V shaped handles of pliers are represented as compositions with unique ids in Table \ref{tab:tools_sci1} and \ref{tab:tools_sci2}, and Table \ref{tab:tools_pliers1} and \ref{tab:tools_pliers2}, respectively.

\begin{table}[t!]
\caption{Results on images of Scissor-1 object belonging to Scissor category obtained from Tools Dataset} \label{tab:tools_sci1}
    \begin{tabular}{|>{\centering\arraybackslash}m{1.8cm}|>{\centering\arraybackslash}m{2.5cm}|>{\centering\arraybackslash}m{2.5cm}|>{\centering\arraybackslash}m{2.5cm}|>{\centering\arraybackslash}m{2.5cm}|}  
  \hline
  \textbf{Object Name}, \textbf{Articulation ID}, \textit{Layer ID} & \multicolumn{2}{|c|}{$l$} & \multicolumn{2}{|c|}{$l+1$} \\
      \hline & Part Realizations $R ^l$ & Object Graph $\mathbb{G}_l$ & Part Realizations $R ^{l+1}$ & Object Graph $\mathbb{G}_{l+1}$ \\
   
      \hline \textbf{Scissor-1, Art-1, $l=6$} &  
      \includegraphics[scale=0.25]{images/tools/01_01_level6-eps-converted-to.pdf} & \includegraphics[scale=0.25]{images/tools/01_01_level6edges-eps-converted-to.pdf} & \includegraphics[scale=0.25]{images/tools/01_01_level7-eps-converted-to.pdf} & \includegraphics[scale=0.25]{images/tools/01_01_level7edges-eps-converted-to.pdf}   \\
      
      \hline \textbf{Scissor-1, Art-2, $l=6$} &  
      \includegraphics[scale=0.25]{images/tools/01_02_level6-eps-converted-to.pdf} & \includegraphics[scale=0.25]{images/tools/01_02_level6edges-eps-converted-to.pdf} & \includegraphics[scale=0.25]{images/tools/01_02_level7-eps-converted-to.pdf} & \includegraphics[scale=0.25]{images/tools/01_02_level7edges-eps-converted-to.pdf}   \\
      
      \hline \textbf{Scissor-1, Art-3, $l=6$} &  
      \includegraphics[scale=0.25]{images/tools/01_03_level6-eps-converted-to.pdf} & \includegraphics[scale=0.25]{images/tools/01_03_level6edges-eps-converted-to.pdf} & \includegraphics[scale=0.25]{images/tools/01_03_level7-eps-converted-to.pdf} & \includegraphics[scale=0.25]{images/tools/01_03_level7edges-eps-converted-to.pdf}   \\
      
      \hline \textbf{Scissor-1, Art-4, $l=5$} &  
      \includegraphics[scale=0.25]{images/tools/01_04_level5-eps-converted-to.pdf} & \includegraphics[scale=0.25]{images/tools/01_04_level5edges-eps-converted-to.pdf} & \includegraphics[scale=0.25]{images/tools/01_04_level6-eps-converted-to.pdf} & \includegraphics[scale=0.25]{images/tools/01_04_level6edges-eps-converted-to.pdf}   \\
      
      \hline \textbf{Scissor-1, Art-5, $l=6$} &  
      \includegraphics[scale=0.25]{images/tools/01_05_level6-eps-converted-to.pdf} & \includegraphics[scale=0.25]{images/tools/01_05_level6edges-eps-converted-to.pdf} & \includegraphics[scale=0.25]{images/tools/01_05_level7-eps-converted-to.pdf} & \includegraphics[scale=0.25]{images/tools/01_05_level7edges-eps-converted-to.pdf}   \\
       
      \hline
  \end{tabular}
  \end{table}

\begin{table}[t!]
\caption{Results on images of Scissor-2 object belonging to Scissor category obtained from Tools Dataset} \label{tab:tools_sci2}
    \begin{tabular}{|>{\centering\arraybackslash}m{1.8cm}|>{\centering\arraybackslash}m{2.5cm}|>{\centering\arraybackslash}m{2.5cm}|>{\centering\arraybackslash}m{2.5cm}|>{\centering\arraybackslash}m{2.5cm}|}  
  \hline
  \textbf{Object Name}, \textbf{Articulation ID}, \textit{Layer ID} & \multicolumn{2}{|c|}{$l$} & \multicolumn{2}{|c|}{$l+1$} \\
      \hline & Part Realizations $R ^l$ & Object Graph $\mathbb{G}_l$ & Part Realizations $R ^{l+1}$ & Object Graph $\mathbb{G}_{l+1}$ \\
      
      \hline \textbf{Scissor-2, Art-1, $l=5$} &  
      \includegraphics[scale=0.25]{images/tools/07_01_level5-eps-converted-to.pdf} & \includegraphics[scale=0.25]{images/tools/07_01_level5edges-eps-converted-to.pdf} & \includegraphics[scale=0.25]{images/tools/07_01_level6-eps-converted-to.pdf} & \includegraphics[scale=0.25]{images/tools/07_01_level6edges-eps-converted-to.pdf}   \\
      
      \hline \textbf{Scissor-2, Art-2, $l=5$} &  
      \includegraphics[scale=0.25]{images/tools/07_02_level5-eps-converted-to.pdf} & \includegraphics[scale=0.25]{images/tools/07_02_level5edges-eps-converted-to.pdf} & \includegraphics[scale=0.25]{images/tools/07_02_level6-eps-converted-to.pdf} & \includegraphics[scale=0.25]{images/tools/07_02_level6edges-eps-converted-to.pdf}   \\
      
      \hline \textbf{Scissor-2, Art-3, $l=5$} &  
      \includegraphics[scale=0.25]{images/tools/07_03_level5-eps-converted-to.pdf} & \includegraphics[scale=0.25]{images/tools/07_03_level5edges-eps-converted-to.pdf} & \includegraphics[scale=0.25]{images/tools/07_03_level6-eps-converted-to.pdf} & \includegraphics[scale=0.25]{images/tools/07_03_level6edges-eps-converted-to.pdf}   \\
      
      \hline \textbf{Scissor-2, Art-4, $l=5$} &  
      \includegraphics[scale=0.25]{images/tools/07_04_level5-eps-converted-to.pdf} & \includegraphics[scale=0.25]{images/tools/07_04_level5edges-eps-converted-to.pdf} & \includegraphics[scale=0.25]{images/tools/07_04_level6-eps-converted-to.pdf} & \includegraphics[scale=0.25]{images/tools/07_04_level6edges-eps-converted-to.pdf}   \\
      
      \hline \textbf{Scissor-2, Art-5, $l=5$} &  
      \includegraphics[scale=0.25]{images/tools/07_05_level5-eps-converted-to.pdf} & \includegraphics[scale=0.25]{images/tools/07_05_level5edges-eps-converted-to.pdf} & \includegraphics[scale=0.25]{images/tools/07_05_level6-eps-converted-to.pdf} & \includegraphics[scale=0.25]{images/tools/07_05_level6edges-eps-converted-to.pdf}   \\

      \hline
  \end{tabular}
  \end{table}

\begin{table}[t!]
\caption{Results on images of Pliers-1 object belonging to Pliers category obtained from Tools Dataset} \label{tab:tools_pliers1}
    \begin{tabular}{|>{\centering\arraybackslash}m{1.8cm}|>{\centering\arraybackslash}m{2.5cm}|>{\centering\arraybackslash}m{2.5cm}|>{\centering\arraybackslash}m{2.5cm}|>{\centering\arraybackslash}m{2.5cm}|}  
  \hline
  \textbf{Object Name}, \textbf{Articulation ID} \textit{Layer ID} & \multicolumn{2}{|c|}{$l$} & \multicolumn{2}{|c|}{$l+1$} \\
      \hline & Part Realizations $R ^l$ & Object Graph $\mathbb{G}_l$ & Part Realizations $R ^{l+1}$ & Object Graph $\mathbb{G}_{l+1}$ \\
   
      \hline \textbf{Pliers-1, Art-1, $l=5$} &  
      \includegraphics[scale=0.25]{images/tools/04_01_level5-eps-converted-to.pdf} & \includegraphics[scale=0.25]{images/tools/04_01_level5edges-eps-converted-to.pdf} & \includegraphics[scale=0.25]{images/tools/04_01_level6-eps-converted-to.pdf} & \includegraphics[scale=0.25]{images/tools/04_01_level6edges-eps-converted-to.pdf}   \\
      
      \hline \textbf{Pliers-1, Art-2, $l=4$} &  
      \includegraphics[scale=0.25]{images/tools/04_02_level4-eps-converted-to.pdf} & \includegraphics[scale=0.25]{images/tools/04_02_level4edges-eps-converted-to.pdf} & \includegraphics[scale=0.25]{images/tools/04_02_level5-eps-converted-to.pdf} & \includegraphics[scale=0.25]{images/tools/04_02_level5edges-eps-converted-to.pdf}   \\
      
      \hline \textbf{Pliers-1, Art-3, $l=5$} &  
      \includegraphics[scale=0.25]{images/tools/04_03_level5-eps-converted-to.pdf} & \includegraphics[scale=0.25]{images/tools/04_03_level5edges-eps-converted-to.pdf} & \includegraphics[scale=0.25]{images/tools/04_03_level6-eps-converted-to.pdf} & \includegraphics[scale=0.25]{images/tools/04_03_level6edges-eps-converted-to.pdf}   \\
      
      \hline \textbf{Pliers-1, Art-4, $l=5$} &  
      \includegraphics[scale=0.25]{images/tools/04_04_level5-eps-converted-to.pdf} & \includegraphics[scale=0.25]{images/tools/04_04_level5edges-eps-converted-to.pdf} & \includegraphics[scale=0.25]{images/tools/04_04_level6-eps-converted-to.pdf} & \includegraphics[scale=0.25]{images/tools/04_04_level6edges-eps-converted-to.pdf}   \\
      
      \hline \textbf{Pliers-1, Art-5, $l=5$} &  
      \includegraphics[scale=0.25]{images/tools/04_05_level5-eps-converted-to.pdf} & \includegraphics[scale=0.25]{images/tools/04_05_level6edges-eps-converted-to.pdf} & \includegraphics[scale=0.25]{images/tools/04_05_level6-eps-converted-to.pdf} & \includegraphics[scale=0.25]{images/tools/04_05_level6edges-eps-converted-to.pdf}   \\
      
      \hline 
      
  \end{tabular}
  \end{table}      
      
\begin{table}[t!]
\caption{Results on images  of Pliers-2 object belonging to Pliers category obtained from Tools Dataset} \label{tab:tools_pliers2}
    \begin{tabular}{|>{\centering\arraybackslash}m{1.8cm}|>{\centering\arraybackslash}m{2.5cm}|>{\centering\arraybackslash}m{2.5cm}|>{\centering\arraybackslash}m{2.5cm}|>{\centering\arraybackslash}m{2.5cm}|}  
  \hline
  \textbf{Object Name}, \textbf{Articulation ID} \textit{Layer ID} & \multicolumn{2}{|c|}{$l$} & \multicolumn{2}{|c|}{$l+1$} \\
      \hline & Part Realizations $R ^l$ & Object Graph $\mathbb{G}_l$ & Part Realizations $R ^{l+1}$ & Object Graph $\mathbb{G}_{l+1}$ \\  
      
      \hline \textbf{Pliers-2, Art-1, $l=5$} &  
      \includegraphics[scale=0.2]{images/tools/06_01_level5-eps-converted-to.pdf} & \includegraphics[scale=0.2]{images/tools/06_01_level5edges-eps-converted-to.pdf} & \includegraphics[scale=0.2]{images/tools/06_01_level6-eps-converted-to.pdf} & \includegraphics[scale=0.2]{images/tools/06_01_level6edges-eps-converted-to.pdf}   \\
      
      \hline \textbf{Pliers-2, Art-2, $l=5$} &  
      \includegraphics[scale=0.2]{images/tools/06_02_level5-eps-converted-to.pdf} & \includegraphics[scale=0.2]{images/tools/06_02_level5edges-eps-converted-to.pdf} & \includegraphics[scale=0.2]{images/tools/06_02_level6-eps-converted-to.pdf} & \includegraphics[scale=0.2]{images/tools/06_02_level6edges-eps-converted-to.pdf}   \\
      
      \hline \textbf{Pliers-2, Art-3, $l=5$} &  
      \includegraphics[scale=0.2]{images/tools/06_03_level5-eps-converted-to.pdf} & \includegraphics[scale=0.2]{images/tools/06_03_level5edges-eps-converted-to.pdf} & \includegraphics[scale=0.2]{images/tools/06_03_level6-eps-converted-to.pdf} & \includegraphics[scale=0.2]{images/tools/06_03_level6edges-eps-converted-to.pdf}   \\
      
      \hline \textbf{Pliers-2, Art-4, $l=5$} &  
      \includegraphics[scale=0.2]{images/tools/06_04_level5-eps-converted-to.pdf} & \includegraphics[scale=0.2]{images/tools/06_04_level5edges-eps-converted-to.pdf} & \includegraphics[scale=0.2]{images/tools/06_04_level6-eps-converted-to.pdf} & \includegraphics[scale=0.2]{images/tools/06_04_level6edges-eps-converted-to.pdf}   \\
      
      \hline \textbf{Pliers-2, Art-5, $l=5$} &  
      \includegraphics[scale=0.2]{images/tools/06_05_level5-eps-converted-to.pdf} & \includegraphics[scale=0.2]{images/tools/06_05_level5edges-eps-converted-to.pdf} & \includegraphics[scale=0.2]{images/tools/06_05_level6-eps-converted-to.pdf} & \includegraphics[scale=0.2]{images/tools/06_05_level6edges-eps-converted-to.pdf}   \\

      \hline
  \end{tabular}
  \end{table}

\newpage




%

\clearpage
\pagebreak

\bibliographystyle{IEEEtranS}
\bibliography{IEEEexample}
